\documentclass[lettersize,journal]{IEEEtran}
\usepackage{amsmath,amsfonts}

\usepackage{array}      % 先加载 array
\usepackage{tabularx}   % 再加载 tabularx
\usepackage{booktabs}
\usepackage{multirow}
\usepackage{multicol}

\usepackage{makecell}      % 用于单元格内换行
\usepackage{xcolor}        % 用于颜色
\usepackage{colortbl}      % 用于表格着色
\usepackage{threeparttable} % 用于表格注释
\usepackage{graphicx}      % 用于resizebox

\usepackage{graphicx}
\usepackage{stfloats}
\usepackage[caption=false,font=normalsize,labelfont=sf,textfont=sf]{subfig}

\usepackage{algorithmic}
\usepackage{algorithm}

\usepackage{textcomp}
\usepackage{url}
\usepackage{verbatim}
\usepackage{cite}
\usepackage{pifont}    % 只加载一次
\usepackage{utfsym}

\usepackage{hyperref}
\usepackage{cleveref}

\usepackage{xcolor}
\usepackage{soul}
\usepackage{cancel}

\UseRawInputEncoding

\usepackage{pdfpages} % 用于整页插入外部PDF

\begin{document}

% \includepdf[pages=-,pagecommand=\thispagestyle{empty}]{respond.pdf}
% \clearpage
% \setcounter{page}{1} % 让正文从第1页开始编号（可按需改）

\title{Whole-Body Integrated Motion Planning \\ for Aerial Manipulators}

% \author{Weiliang Deng$^{1*}$, Hongming Chen$^{1*}$, Biyu Ye$^{1}$, Haoran Chen$^{1}$, Ziliang Li$^{1}$ and Ximin Lyu$^{1,2\dagger}$

% \thanks{%
% \sloppy
% \raggedright
% ${}^{*}$ \textbf{Equal contribution}.
% ${}^{\mathbf{\dagger}}$ \textbf{Corresponding Author}.\linebreak
% ${\ \ \ }^1$ School of Intelligent Systems Engineering, Sun Yat-sen University, Guangzhou, 510275, China.${\ \ }^2$ Differential Robotics Technology Co., Ltd., Hangzhou, China.
% Emails: \{dengwliang, chenhm223, yeby9, chenhr66, litleong3\}@mail2.sysu.edu.cn, lvxm6@mail.sysu.edu.cn%
% }
% }

\author{Weiliang Deng, Hongming Chen, Biyu Ye, Haoran Chen, Ziliang Li and Ximin Lyu
\thanks{Manuscript received: January 15, 2025; Revised June 28, 2025; Accepted September 26, 2025.}
\thanks{This paper was recommended for publication by Editor Paolo Robuffo Giordano upon evaluation of the Associate Editor and Reviewers’ comments. This work is supported by the National Key Research and Development Program of China (Grant No. 2023YFB4706600), the National Natural Science Foundation of China (Grant No. 62303495), the Guangdong-Hongkong-Macao Joint Research of Science and Technology Planning Funding from Guangdong Province (Grant No. 2023A0505010019), and the Young Talent Support Project of Guangzhou Association for Science and Technology (Grant No. QT-2025-004). (\textit{Weiliang Deng and Hongming Chen contributed equally to this work.)  (Corresponding authors: Ximin Lyu.)}}
\thanks{
Weiliang Deng, Hongming Chen, Biyu Ye, Haoran Chen, Ziliang Li, and Ximin Lyu are with the School of Intelligent Systems Engineering, Sun Yat-sen University, Guangzhou, 510275, China (e-mail: \{dengwliang, chenhm223, yeby9, chenhr66, litleong3\}@mail2.sysu.edu.cn, lvxm6@mail.sysu.edu.cn). Ximin Lyu is also with Differential Robotics Technology Co., Ltd., Hangzhou, China.
}
}


% {Shell \MakeLowercase{\textit{\textit{et al.}}}: A Sample Article Using IEEEtran.cls for IEEE Journals}


\maketitle

\begin{abstract}
Expressive motion planning for Aerial Manipulators (AMs) is essential for tackling complex manipulation tasks, yet achieving coupled trajectory planning adaptive to various tasks remains challenging, especially for those requiring aggressive maneuvers. In this work, we propose a novel whole-body integrated motion planning framework for quadrotor-based AMs that leverages flexible waypoint constraints to achieve versatile manipulation capabilities.
These waypoint constraints enable the specification of individual position requirements for either the quadrotor or end-effector, while also accommodating higher-order velocity and orientation constraints for complex manipulation tasks.
To implement our framework, we exploit spatio-temporal trajectory characteristics and formulate an optimization problem to generate feasible trajectories for both the quadrotor and manipulator while ensuring collision avoidance considering varying robot configurations, dynamic feasibility, and kinematic feasibility. Furthermore, to enhance the maneuverability for specific tasks, we employ Imitation Learning (IL) to facilitate the optimization process to avoid poor local optima. The effectiveness of our framework is validated through comprehensive simulations and real-world experiments, where we successfully demonstrate nine fundamental manipulation skills across various environments.
\end{abstract}

% \def\abstractname{Note to Practitioners}
% \begin{abstract}
% This paper presents a whole-body integrated motion planning framework for aerial manipulators, enabling collision-free navigation through complex environments while satisfying feasibility and multiple waypoint constraints.
% Our implementation demonstrates the framework on delta-type aerial manipulators, and the proposed dynamic ellipsoid approximation concept is generalizable to other deformable structures, enabling broad applicability across diverse industrial scenarios.
% Extensive experiments in different practical applications validate the framework's reliability and real-world performance.
% \end{abstract}


\section{Introduction}
% Background
\IEEEPARstart{A}{erial} manipulators, which consist of Unmanned Aerial Vehicles (UAVs) and manipulators, have garnered significant interest from researchers and industries due to their wide operational range and agile manipulation capabilities \cite{ollero_past_2022}. These systems have proven to operate effectively in challenging or hazardous environments, facilitating applications across various fields such as 3D printing, inspection, transportation, grasping, repair and perching \cite{zhang_aerial_2022, trujillo_novel_2019, luo_time-optimal_2023, chermprayong_integrated_2019, hang_perching_2019, nguyen_novel_2018}.
% What is the problem? 
Various aspects of aerial manipulators have been extensively studied, including platform design \cite{xu_biomimetic_2024, thomas_avian-inspired_2013, ryll_6d_2019, bodie_active_2021}, motion planning \cite{cao_motion_2024, lee_planning_2018}, and precise control \cite{wang_millimeter-level_2024, chen_adaptive_2023, cao_eso-based_nodate}. However, most existing approaches have been developed for specific tasks with corresponding algorithms and hardware \cite{he_flying_2025}. Developing integrated planning frameworks that can autonomously accomplish multiple manipulation skills, particularly those requiring aggressive maneuvers, remains challenging.

\begin{figure}[t]
\centering
\includegraphics[width=1\linewidth]{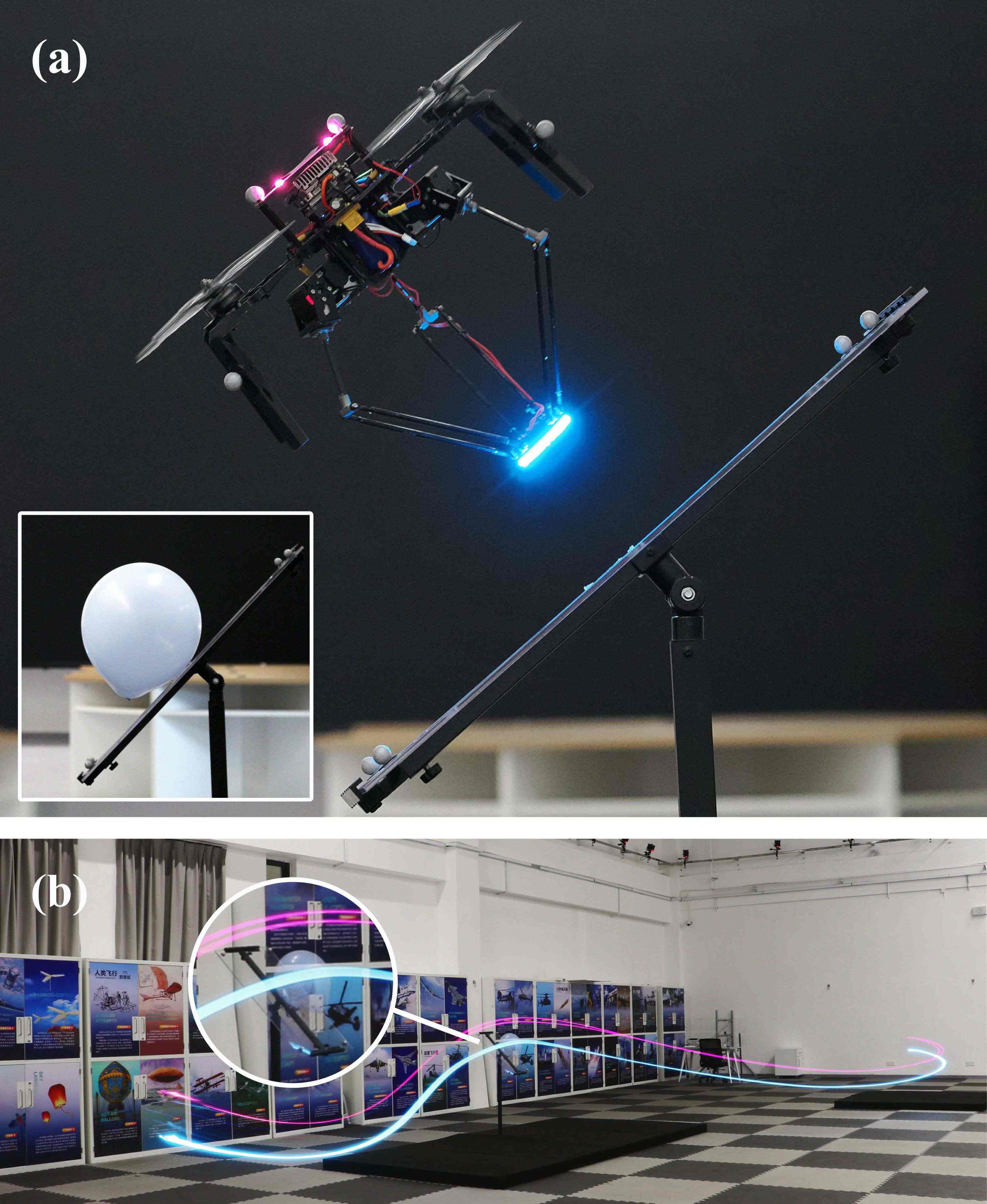}
\caption{Demonstration of aerial striking using whole-body integrated motion planning. (a) The aerial manipulator executes a planned trajectory to reach a constrained waypoint, maintaining an attitude parallel to the inclined plane while configuring its delta arm for balloon striking. (b) Time-lapse visualization capturing the complete execution of the planned motion. 
}
\label{fig:cover}
\end{figure}

% Why is it important? 
Unlike decoupled approaches that separately plan UAV and manipulator trajectories, coupled planning enables coordinated motion generation for the entire system, preserving its full maneuverability for complex tasks \cite{spahn_coupled_2021}. 
% In developing such planning frameworks, collision avoidance emerges as a critical consideration for safe environmental interaction, which has been overlooked in some previous work \cite{spica_aerial_2012, luo_time-optimal_2023}. This capability is essential for traversing complex environments while guaranteeing collision-free manipulation operations.
%%%%%%%%%%%%%%%%%   couple planning framework  %%%%%%%%%%%%%%%%%%%
% Why is the problem hard? What makes it challenging?
However, coupled planning for aerial manipulators presents significant challenges, with the most fundamental stemming from the high-dimensional state space. The inclusion of additional joints from the manipulators greatly increases the system's dimensionality, resulting in a vastly expanded search space.
% How far has existing work come? Why hasn't the problem been solved? What is the stumbling block? What is the frontier?
To address the challenges of high-dimensional motion planning, various approaches have been developed. 
Early work proposes an optimization-based framework for aerial manipulators, focusing on aerial grasping of moving targets \cite{spica_aerial_2012}. However, this approach primarily prioritizes time minimization while overlooking other critical performance metrics, such as energy consumption.
The method in \cite{tognon_control-aware_2018} performs trajectory search in the task space and employs a controller as a local planner to generate whole-body motions. However, this approach lacks global optimization considerations and therefore cannot guarantee overall trajectory smoothness or energy efficiency.
The approach in \cite{suseong_kim_aerial_2013} integrates end-effector and UAV dynamics within an NMPC framework for aerial writing tasks. 
More recently, a discrete-mechanics-based time-optimal trajectory planning framework with complementarity constraints was developed for aerial manipulators \cite{luo_time-optimal_2023}.
However, both approaches are designed for specific single tasks and their formulations make them difficult to systematically determine more constrained intermediate states, for example striking a point or being inclined at a certain angle, and lack flexibility to address broader manipulation tasks.

\begin{figure}
    \centering
    \includegraphics[width=1\linewidth]{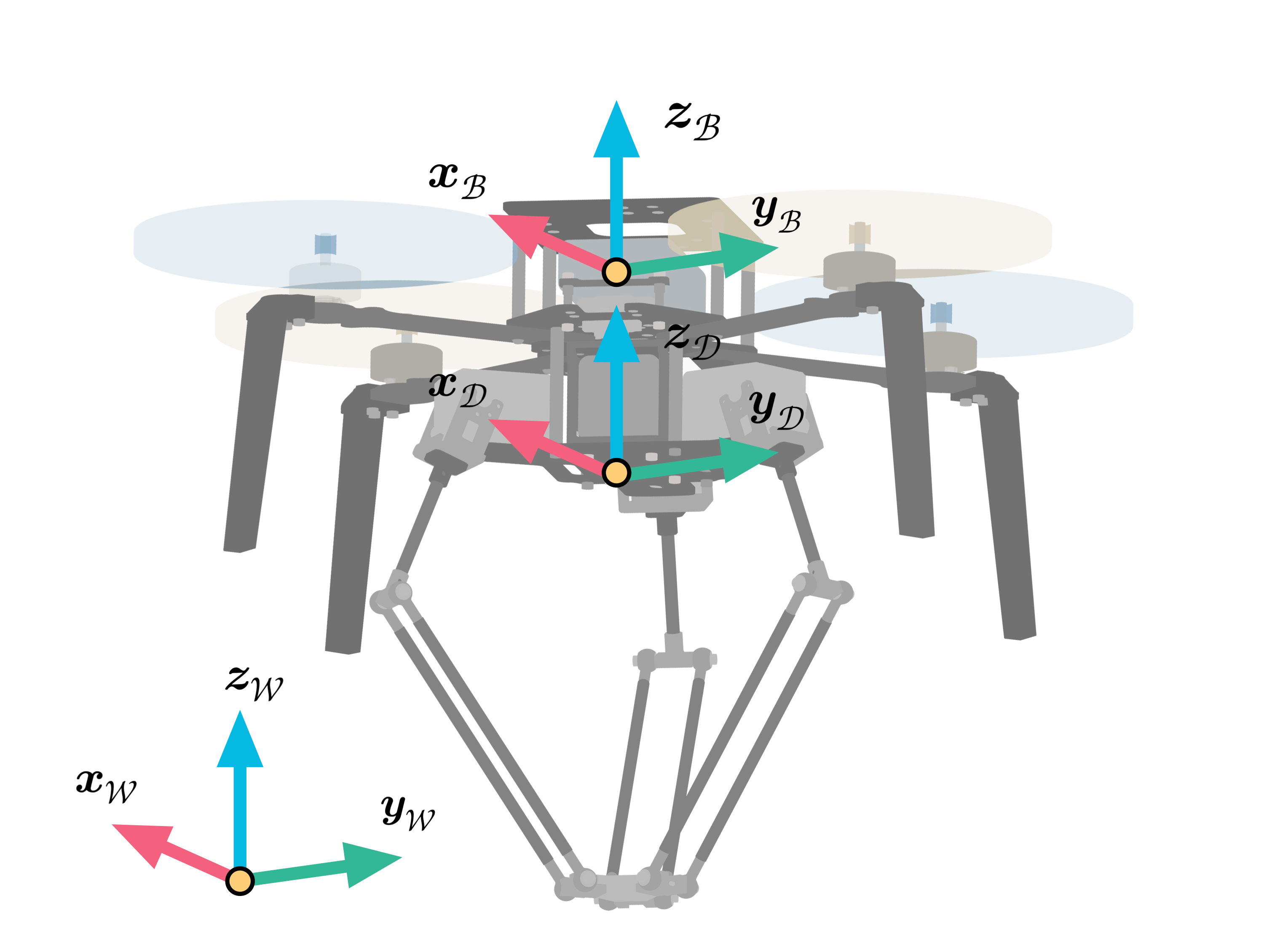}
    \caption{Coordinate frames of the aerial manipulator system: World frame ($\mathcal{F}_\mathcal{W}$), Body frame ($\mathcal{F}_\mathcal{B}$), and Delta frame ($\mathcal{F}_\mathcal{D}$).}
    \label{fig:Coordinate}
\end{figure}

To enable versatile aerial manipulation, a planning framework should be capable of accommodating various simple skills, including grasping, striking, pushing, pulling, writing, and so on\cite{brunner_planning-and-control_2022, guo_flying_2025}. Intuitively, these complex skills can be naturally broken down into fundamental subtasks: point traversing for the quadrotor, point reaching for the end-effector, and combining with extra velocity and orientation requirements as needed at these points.
Therefore, we introduce flexible partial waypoint constraints, which allow selective specification of position, velocity, or orientation requirements for individual quadrotors or end-effectors at different waypoints, rather than imposing complete state constraints at every point.
However, the diverse types of waypoints pose significant challenges due to the need to satisfy precise intermediate-state requirements while maintaining feasibility across the entire trajectory \cite{foehn_time-optimal_2021}.
Existing methods for handling waypoint constraints fall into two main categories: multi-stage sampling and optimization integration methods. Multi-stage sampling divides the trajectory into segments by setting multiple start and goal points \cite{kim_sampling-based_nodate}. While this approach can effectively handle partial waypoint constraint problems, it primarily focuses on position without considering orientation information. Introducing orientation constraints may lead to a significantly enlarged search space or even result in the inability to find feasible solutions.
For optimization integration methods, several approaches have been proposed with varying strategies. The method in \cite{seo_locally_2017} treats both polynomial trajectories and waypoint arrival times as optimization variables, then penalizes the distance between the end-effector and target points at those specific times. This method provides valuable insights, but for multiple waypoint constraints, it should introduce more time optimization variables to determine the waypoint constraints scope and more variable constraints which may make it difficult to solve. Gaussian penalty functions have been used as soft constraints to incorporate waypoint constraints into MPC \cite{neunert_fast_2016}. However, since arrival times are provided by higher-level planners, this approach still lacks adaptability and an inappropriate time may lead to an unfeasible solution.
Recently, a general spatio-temporal optimization framework with geometric constraints, named MINCO \cite{wang_geometrically_2022}, has been proposed to address whole-body trajectory generation for quadrotors. It operates by optimizing the positions of intermediate waypoints and the intervals between these points and then solves the complete trajectory. This optimization structure makes it particularly well-suited for incorporating waypoint constraints on these intermediate points, which forms the foundation for our implementation approach.

% IL加一个图
For most skills, given a target waypoint along with some orientation or velocity constraints, it is generally possible to optimize a feasible trajectory that accomplishes the task. 
However, for highly dynamic maneuvers, such as parallel grasping or striking on an inclined plane, as shown in \Cref{fig:cover} (a), directly using a single waypoint constraint may converge to poor suboptimal solutions. 
For such highly attitude-constrained trajectories, there exist inherent underlying motion patterns that should be leveraged. For example, quadrotors tend to follow arc-shaped trajectory patterns to satisfy their dynamics, as shown in \Cref{fig:cover} (b), with this tendency becoming more pronounced as the attitude angle increases.
Therefore, we propose a novel guided optimization approach for aerial manipulations that initially imposes additional waypoint constraints to steer the trajectory optimization toward desired local optima that better satisfy both task requirements and dynamic feasibility constraints, and then progressively relaxes these auxiliary constraints to allow for further refinement.
The question then becomes: how should we properly select these waypoints?
Recently, IL has demonstrated significant capabilities in the robot manipulation domain. For instance, Diffusion Policy \cite{chi_diffusion_2024} learns data distributions through imitation learning to achieve policy learning, while OpenVLA \cite{kim_openvla_2024} and $\pi_0$ \cite{black__0_2024} achieve task generalization through imitation learning on large-scale robot training data. 
A recent work, Flying Hand \cite{he_flying_2025}, is the first to incorporate teleoperation into aerial manipulators, training an ACT \cite{zhao_learning_2023} policy and achieving some basic tasks. 
Current mainstream IL approaches typically learn to output single-step actions or complete trajectories based on environmental information. 
However, we aim to leverage this capability of imitation learning through a lightweight network that imitates our aerial manipulator's local behavior near target points when performing large-attitude manipulation, using this information as a prior for guided optimization.

% What does our paper contribute? What is the key idea? What is the magic trick? What is the new insight or technique that enables us to advance the frontier?
% First, you need to clearly state what it is that the paper does.
% Second, it is often beneficial to highlight a key idea, a central insight, something that allows you to do what you do.
In this paper, we propose a whole-body integrated planning framework for aerial manipulators that incorporates flexible waypoint constraints which can be used to freely combine into many fundamental skills.
First, we develop an improved safe flight corridor (SFC) generation strategy that not only expands the feasible solution space but also guarantees the feasible waypoint constraint implementation.
Second, we modify the dimension and the dynamics from \cite{wang_geometrically_2022} and then formulate a spatio-temporal optimization problem for aerial manipulators to generate a 6-dimensional trajectory, which includes the quadrotor's trajectory in the world frame and the manipulator's trajectory in the local frame. In addition, we improve the constraint formulation from \cite{wang_geometrically_2022}, specifically by introducing a varying ellipsoid to model the changing collision volume of the aerial manipulator caused by the motion of the manipulator.
Third, we introduce a comprehensive approach for handling waypoint constraints and incorporate various task-specific constraints, including partial or complete position constraints (with individual $x$, $y$, $z$ axis constraint), velocity, and orientation requirements.
Building upon this constraint handling approach, we further introduce the concept of IL-guided optimization that employs IL to generate guide points, which serve as waypoint constraints to improve trajectory convergence.
Finally, we validate our framework through comprehensive comparative and ablation studies in simulation, as well as real-world aerial manipulation experiments to demonstrate its effectiveness. The main contributions of this work are:
\begin{enumerate}
    \item \textbf{Whole-body Planning Framework}: We propose a novel whole-body integrated motion planning framework for aerial manipulators. This framework simultaneously optimizes the trajectory of both the quadrotor and the manipulator to achieve coordinating whole-body motion and complete a series of tasks, strike, grasp, lift, write, pull, push, wind, cross and press. Additionally, we develop a novel varying ellipsoid method to model the changing collision volume to guarantee the whole-body collision-free manipulations under any requirements. 

    \item \textbf{Flexible Partial Waypoint Constraint}: We introduce a comprehensive waypoint constraint formulation that enables selective specification of position, velocity, or orientation requirements for individual quadrotors or end-effectors at different waypoints. This approach allows flexible task-specific partial waypoint constraints that only constrain the necessary degrees of freedom while leaving others free for optimization, providing significantly greater flexibility and more constraint options compared to existing methods.

    \item \textbf{IL-Guided Optimization for Complex Tasks}: We introduce an IL-guided optimization approach to overcome poor local optima in challenging manipulation scenarios, such as grasping or striking on inclined surfaces. The method learns from high-quality demonstration data to generate a sequence of waypoints to guide the optimization process in the first stage, and then progressively relaxes these guided constraints in the second stage to find better solutions.

    % \item \textbf{Comprehensive Validation and Open-source Implementation:} We demonstrate the versatility and effectiveness of our framework through extensive validation, including diverse benchmark tasks, ablation studies, simulations on different types of manipulators, and real-world aerial manipulation experiments. The complete framework is made publicly available to support the research community and promote reproducible research in aerial manipulation planning. The source code is available at: \url{https://github.com/SYSU-HILAB/am-planner.git}.
    \item \textbf{Comprehensive Validation and Open-source Implementation:} We demonstrate the versatility and effectiveness of our framework through extensive validation, including diverse benchmark tasks, ablation studies, simulations on different types of manipulators, and real-world aerial manipulation experiments. The complete framework is made publicly available to support the research community and promote reproducible research in aerial manipulation planning.\footnote{\url{https://github.com/SYSU-HILAB/am-planner.git}}
\end{enumerate}

\begin{figure*}
\centering
\includegraphics[width=1\linewidth]{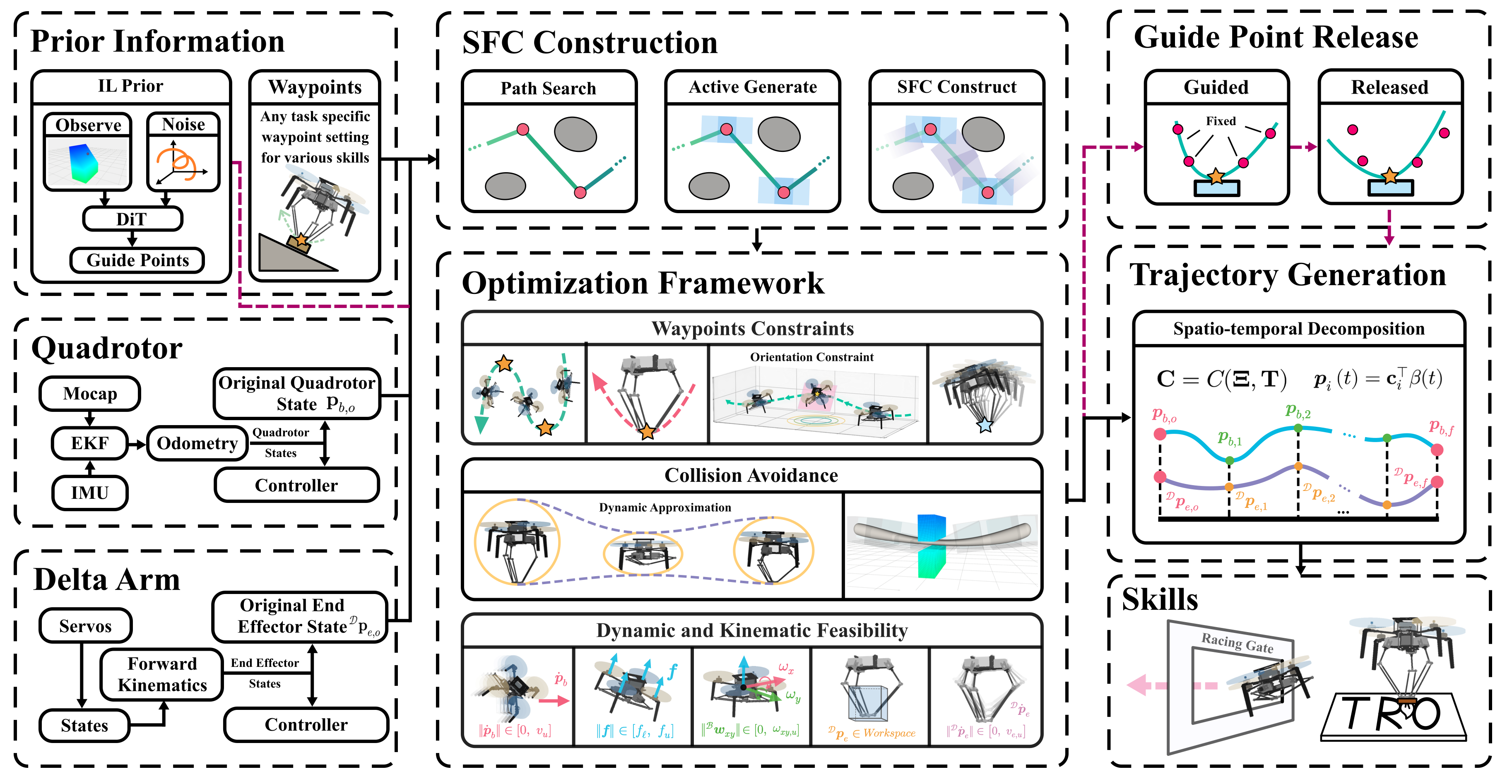}
\caption{Overview of the proposed integrated motion planning framework for aerial manipulation. The framework consists of seven key components: (a) Prior Information module that employs a Diffusion Transformer to generate trajectory tendencies and processes user-defined waypoint constraints; (b) Quadrotor module integrating motion capture (Mocap), Extended Kalman Filter (EKF), and IMU for robust state estimation; (c) Delta Arm module handling servo control and forward kinematics for Cartesian space state computation; (d) Safe Flight Corridor (SFC) Construction module for path searching, active polyhedron generation, and conventional SFC construction; (e) Optimization Framework addressing waypoint constraints, collision avoidance, and dynamic/kinematic feasibility; (f) IL-guided two-stage optimization where guide points serve as hard constraints in the initial stage and are subsequently released for free optimization; (g) Trajectory Generation module producing smooth polynomial trajectories; and (h) Skills module performing various basic manipulation skills.}
\label{fig:overview}
\end{figure*}

Our framework overview is depicted in \Cref{fig:overview}. The paper is organized as follows. \Cref{sec: preliminaries} introduces the notation, system model, and trajectory representation. \Cref{sec: active sfc generation} presents the improved SFC generation strategy, while \Cref{sec: Problem Formulation} elaborates on the basic optimization problem formulation. \Cref{sec: Waypoint_Constraint} introduces the implementation of waypoint constraint and IL-guided optimization.
\Cref{sec: experiments} validates the effectiveness of the proposed method through simulations and experiments. Finally, conclusions are presented in \Cref{sec: conclusion}.

\section{Preliminaries}\label{sec: preliminaries}
\subsection{Notation}
The aerial manipulator system discussed in this paper consists of a quadrotor platform integrated with a 3-DOF parallel delta robot with 3-RSS configuration similar to \cite{mansoor_accuracy_2018}, as illustrated in \Cref{fig:Coordinate}. We experimentally determine the Center of Mass (CoM) of the entire aerial manipulator system by placing the delta arm in its initial state and measuring the resulting CoM, which we denote as $\boldsymbol{c}_{m0}$.
For convenience, we define three distinct coordinate frames:

\begin{enumerate}
    \item \textbf{World Frame ($\mathcal{F}_\mathcal{W}$)}: This frame is denoted by $\{ \boldsymbol{x}_\mathcal{W}, \boldsymbol{y}_\mathcal{W}, \boldsymbol{z}_\mathcal{W}\}$, where $\boldsymbol{z}_\mathcal{W}$ points upwards, opposing the direction of gravity.
    \item \textbf{Body Frame ($\mathcal{F}_\mathcal{B}$)}: This frame is represented by $\{\boldsymbol{x}_\mathcal{B}, \boldsymbol{y}_\mathcal{B}, \boldsymbol{z}_\mathcal{B}\}$. The origin of this frame is located at $\boldsymbol{c}_{m0}$. In this frame, $\boldsymbol{x}_\mathcal{B}$ points forward and $\boldsymbol{z}_\mathcal{B}$ aligns with the total thrust generated by the rotors.
    \item \textbf{Delta Frame ($\mathcal{F}_\mathcal{D}$)}: This frame is specified as $\{\boldsymbol{x}_\mathcal{D}, \boldsymbol{y}_\mathcal{D}, \boldsymbol{z}_\mathcal{D}\}$ and is related to the body frame $\mathcal{F}_\mathcal{B}$ through a translational transformation only.
\end{enumerate}

Unless explicitly stated otherwise, the pre-superscript of a vector indicates the frame in which the vector is expressed. For conciseness, when a vector is expressed in the world frame, the pre-superscript will be omitted. For instance, $\boldsymbol{p}_b$ represents the position of the CoM of the aerial manipulator, in $\mathcal{F}_\mathcal{W}$, while ${}^{\mathcal{D}}\boldsymbol{p}_e$ denotes the position of the end-effector in $\mathcal{F}_\mathcal{D}$.  Analogously, the matrix $\mathbf{R}_\mathcal{B}$ represents the rotation from  $\mathcal{F}_{\mathcal{W}}$ to $\mathcal{F}_\mathcal{B}$.

Let $\boldsymbol{p}_o = \left[ \boldsymbol{p}_{b,o}^\top, {}^{\mathcal{D}}\boldsymbol{p}_{e,o}^\top \right]^\top \in \mathbb{R}^6$ and $\boldsymbol{p}_f =  \left[ \boldsymbol{p}_{b,f}^\top, {}^{\mathcal{D}}\boldsymbol{p}_{e,f}^\top \right]^\top\in \mathbb{R}^6$ denote the origin and final positions of the aerial manipulator in $\mathcal{F}_\mathcal{W}$ and end-effector in $\mathcal{F}_\mathcal{D}$, respectively. 
We define a waypoint constraint set in frame $\mathcal{F}_\mathcal{W}$ that consists of arbitrary combinations of two types of waypoint constraints:
\begin{align}
\mathbf{W}_{\mathrm{cons}} &:= \{ \boldsymbol{w}_{\lambda} \mid \lambda \in \{1, \ldots, n_\Lambda\} \}\label{eq: wpconstraint}
\end{align}
The two types of waypoint constraints are defined as:
\begin{align}
\boldsymbol{w}_{\lambda} &:= \begin{cases}
\boldsymbol{p}_{b, \lambda}^{\mathrm{cons}}, & \text{if } \lambda \in \Lambda_b \\
\boldsymbol{p}_{e, \lambda}^{\mathrm{cons}}, & \text{if } \lambda \in \Lambda_e
\end{cases}
\end{align}
where $\Lambda_b$ and $\Lambda_e$ are disjoint index sets with $\Lambda_b \cup \Lambda_e = \{1, \ldots, n_\Lambda\}$, representing quadrotor and end-effector waypoint constraints, respectively.

% As we mentioned previously, we have two types of waypoint constraints to choose from. Therefore, we express the waypoint constraint in frame $\mathcal{F}_\mathcal{W}$ for the quadrotor as $\boldsymbol{p}_{b,\lambda}^{\mathrm{cons}}$ and for the end-effector as $\boldsymbol{p}_{e,\lambda}^{\mathrm{cons}}$, where $\lambda \in \{1,2,\ldots,n_\Lambda\}$ denotes the index of the waypoint constraint. 
% We define:
% \begin{align}
% \mathbf{W}_{\mathrm{cons}} &:= \{ \boldsymbol{w}_{\lambda} \mid \lambda \in \{1, \ldots, n_\Lambda\} \} \\
% \boldsymbol{w}_{\lambda} &:= \begin{cases}
% \boldsymbol{p}_{b, \lambda}^{\mathrm{cons}} & \text{if } \lambda \in \Lambda_b \\
% \boldsymbol{p}_{e, \lambda}^{\mathrm{cons}} & \text{if } \lambda \in \Lambda_e
% \end{cases}
% \end{align}
% where $\Lambda_b$ and $\Lambda_e$ are disjoint index sets such that $\Lambda_b \cup \Lambda_e = \{1, \ldots, n_\Lambda\}$, representing quadrotor and end-effector constrained waypoints, respectively.
% Moreover, for aerial manipulators with delta arms, the end-effector and quadrotor share the same orientation. Therefore, we denote $\boldsymbol{n}_\lambda^{\mathrm{cons}}$ as the orientation at the $\lambda$-th waypoint.

% Previous research \cite{seo_locally_2017} leverages the differential flatness of the multirotor and design a robust controller for the entire system to effectively reject the inertial effects induced by the manipulator. Similarly, a previous work \cite{chen_ndob-based_nodate} proposes a controller that compensates for the coupled dynamics between the quadrotor and the delta arm, enabling independent trajectory tracking for both subsystems. Inspired by these works, we preserve the differential flatness property of the quadrotor and employ a similar controller to compensate for the coupled dynamics. Specifically, we first analyze the dynamics of the quadrotor and the delta arm separately, and then combine them into the whole-body dynamics of the aerial manipulator.

\subsection{Modeling}
We use Newton-Euler equations to model the whole-body dynamics. Following \cite{tzoumanikas_aerial_202}, we treat the aerial manipulator as a rigid body system and focus on quasi-static forces from manipulator motion and environmental interactions. This approach is reasonable as the delta arm's mass represents only a small fraction of the total system mass \cite{chen_ndob-based_nodate}, and we will strictly limit the manipulator's velocity during planning. For instantaneous actions such as grasping and striking, we cannot model these transient disturbances, but we do model the environmental  external forces, that occur during sustained operations like pulling and holding grasped objects.

1) \textbf{Aerial Manipulator Dynamics}: 
The coupled dynamics of the aerial manipulator system are described by:
\begin{align}
\dot{\boldsymbol{p}}_{b} &= \boldsymbol{v}_{b}\\
\dot{\boldsymbol{v}}_{b} &=  \frac{1}{m_{c}} \mathbf{R}_\mathcal{B} ( {}^{\mathcal{B}} \boldsymbol{f} + {}^{\mathcal{B}}\boldsymbol{f}_e) - g\boldsymbol{z}_{\mathcal{W}}\\
{}^{\mathcal{B}}\dot{\boldsymbol{\omega}}_{b} &=\mathbf{I}_{c}^{-1}( {}^{\mathcal{B}}\boldsymbol{\tau} + {}^{\mathcal{B}}\boldsymbol{\tau}_e  - {}^{\mathcal{B}}\boldsymbol{\omega}_{b} \times\mathbf{I}_{c} \cdot {}^{\mathcal{B}}\boldsymbol{\omega}_{b})\\
\mathbf{I}_{c} &= \mathbf{I} + m_{e}\cdot\mathrm{diag}( {}^{\mathcal{B}}\boldsymbol{p}_{e} - {}^{\mathcal{B}}\boldsymbol{p}_{e0} )^2
\end{align}
where $\boldsymbol{p}_b$ and $\boldsymbol{v}_b$ represent the position and linear velocity of the aerial manipulator, respectively. The combined system mass and the end-effector mass are denoted by $m_c$ and $m_e$ respectively.
${}^{\mathcal{B}}\boldsymbol{f}$ and ${}^{\mathcal{B}}\boldsymbol{\tau}$ are the thrust and torque generated by the quadrotor, while ${}^{\mathcal{B}}\boldsymbol{f}_e$ and ${}^{\mathcal{B}}\boldsymbol{\tau}_e$ are the force and moment generated by the end-effector.
$\mathbf{I}_c$ is the combined inertia, where $\mathbf{I} = \mathrm{diag}(I_x, I_y, I_z)$ is the inertia tensor of the aerial manipulator with the arm at its initial state. Here, ${}^{\mathcal{B}}\boldsymbol{p}_{e0}$ and ${}^{\mathcal{B}}\boldsymbol{p}_{e}$ denote the initial and current positions of the end-effector in the body frame $\mathcal{F}_\mathcal{B}$, respectively.

Now we consider the force ${}^{\mathcal{B}}\boldsymbol{f}_e$ and moment ${}^{\mathcal{B}}\boldsymbol{\tau}_e$ caused by the end-effector:
\begin{align}
 {}^{\mathcal{B}}\boldsymbol{f}_e &= {}^{\mathcal{B}}\mathbf{R}_{\mathcal{W}}\boldsymbol{f}_{\mathrm{ext}}\\
 {}^{\mathcal{B}}\boldsymbol{\tau}_e &= {}^{\mathcal{B}}\boldsymbol{p}_e \times {}^{\mathcal{B}}\boldsymbol{f}_e  + ({}^{\mathcal{B}}\boldsymbol{p}_e - {}^{\mathcal{B}}\boldsymbol{p}_{e0}) \times ({}^{\mathcal{B}}\mathbf{R}_\mathcal{W} \cdot m_e g\boldsymbol{z}_\mathcal{W})
\end{align}
where $\boldsymbol{f}_{\mathrm{ext}}$ is the external forces applied to the end-effector, for example, the gravitational force when grasping an object, or the frictional force when pulling or pushing.

2) \textbf{Delta arm Kinematics}: 
Our quadrotor incorporates a 3-DOF delta arm mounted directly beneath the vehicle. This configuration offers significant advantages in terms of lightweight construction, cost-effectiveness, and operational speed, making it widely adopted in recent research \cite{zhang_aerial_2022, chen_ndob-based_nodate}. The kinematic model for the delta arm is formulated as follows:
\begin{align}
\boldsymbol{p}_e &= \mathbf{R}_\mathcal{B}({}^\mathcal{B}\boldsymbol{p}_{\mathcal{D}} + {}^\mathcal{D}\boldsymbol{p}_e ) + \boldsymbol{p}_b \label{eq: ee_world_representation}\\
{\boldsymbol{v}_e} &= [\boldsymbol{\omega}_\mathcal{B}]^\times \mathbf{R}_\mathcal{B}({}^\mathcal{B}\boldsymbol{p}_{\mathcal{D}} + {}^\mathcal{D}\boldsymbol{p}_e ) + \mathbf{R}_\mathcal{D} \;^{\mathcal{D}}\boldsymbol{v}_e+
\boldsymbol{v}_b  
\end{align}
where $\boldsymbol{p}_e$ and $\boldsymbol{v}_e$ represent the position and velocity of the end-effector in frame $\mathcal{F}_\mathcal{W}$, and $[\cdot]^\times$ denotes the skew-symmetric matrix operator.

Since the above representation operates in Cartesian space, inverse kinematics (IK) must be employed to establish the mapping between the end-effector position ${}^{\mathcal{D}}\boldsymbol{p}_e$ and the joint angles $\boldsymbol{q} = \left[ q_1, q_2, q_3 \right]^\top \in \mathbb{R}^3$. In this formulation, $L_{u}$ and $L_{l}$ denote the lengths of the upper arm and lower arm, respectively, while $r_s$ represents the circumradius of the static platform and $r_d$ represents the circumradius of the end-effector.
Following the methodology outlined in \cite{Codourey_proceedings_1996}, the relationship between the end-effector position in the delta frame ${}^{\mathcal{D}}\boldsymbol{p}_e$ and the joint angles $\boldsymbol{q}$ is obtained by solving the following equation: 
\begin{equation}
\left\| {}^{\mathcal{D}}\boldsymbol{p}_e - {}^{\mathcal{D}}\mathbf{R}_i \cdot 
\begin{bmatrix}
r_s + L_u \sin q_i \\
0 \\
- L_u \cos q_i \\
\end{bmatrix} \right\|^2_2
= L_l^2,
\end{equation}
where $i \in \{1,2,3\}$, and ${}^{\mathcal{D}}\mathbf{R}_i$ denotes the $z$-axis rotation matrix by $\{ 0, \frac{2}{3}\pi, \frac{4}{3}\pi \}$ radians, respectively.
By differentiating both sides of the above equation with respect to time and rearranging the terms, we obtain:
\begin{equation}
\dot{\boldsymbol{q}}=\mathbf{J} \ {}^{\mathcal{D}}\dot{\boldsymbol{p}}_e,
\end{equation}
where $\mathbf{J}$ denotes the Jacobian matrix. The explicit expression for this matrix is detailed in \cite{Codourey_proceedings_1996}.

\subsection{Trajectory Representation}\label{sec: trajectory_representation}
In this study, we adopt a hybrid coordinate representation for trajectory planning: the quadrotor's trajectory is planned in the frame $\mathcal{F}_\mathcal{W}$ while the end-effector trajectory is planned in the frame $\mathcal{F}_\mathcal{D}$. This definition offers two key advantages: (1) it enables better capture of the end-effector's motion relative to the quadrotor, facilitating accurate collision volume modeling and workspace constraint enforcement; (2) when no manipulation tasks are required, the system can directly operate as a standard quadrotor without representation switching.

Based on this design, we utilize 6-dimensional spatio-temporal polynomial trajectories that decompose multi-segment trajectories into waypoints and their corresponding time intervals. We denote the trajectory as:
\begin{equation}\label{eq:minco}
\mathfrak{T} = \left\{\boldsymbol{p}(t):\left[0, T_{i}\right] \mapsto \mathbb{R}^m \mid \mathbf{C}=C(\mathbf{q}, \mathbf{T}),\; i = 1,\ldots, M\right\}
\end{equation}
where $m = 6$ represents the dimension of the trajectory.
$\mathbf{C}=\left[\mathbf{c}_1^\top, \cdots, \mathbf{c}_M^\top\right]^\top \in \mathbb{R}^{2Ms \times 6}$ denotes the polynomial coefficient matrix, where $M$ represents the number of segments in the polynomial trajectory and $s$ represents the order of the integrator chain for control effort.
$\mathbf{q}=\left[\boldsymbol{q}_1, \cdots, \boldsymbol{q}_{M-1}\right]^\top\in \mathbb{R}^{(M-1)\times 6}$ represents the intermediate points, where the first 3 dimensions correspond to the quadrotor and the last 3 dimensions correspond to the end-effector.
$\mathbf{T}=\left[T_1, \cdots, T_M\right]^\top \in \mathbb{R}_{>0}^M$ represents the non-negative time vector.
$C(\mathbf{q}, \mathbf{T})$ denotes the parameter mapping constructed from Theorem 2 in \cite{wang_geometrically_2022}.

For the $i$-th segment of the trajectory, the position at time $t$ is defined as:
\begin{equation}\label{eq: p=cbeta}
\boldsymbol{p}_i(t) = \mathbf{c}_i^\top \boldsymbol{\beta}(t), \quad t \in \left[0, T_i\right],
\end{equation}
where $i \in \{1, 2, \ldots, M\}$, and $T_i$ represents the duration of the $i$-th segment. The polynomial coefficient matrix for the $i$-th segment is denoted by  $\mathbf{c}_i \in \mathbb{R}^{2s \times 6}$, and the basis vector function $\boldsymbol{\beta}(x)$ is defined as $\boldsymbol{\beta}(x) = \left[1, x, \cdots, x^{2s-1}\right]^\top \in \mathbb{R}^{2s}$.

\begin{algorithm}[htbp]
\caption{Active SFC Generation}
\label{alg:sfc_generation}
\begin{algorithmic}[1]
\STATE \textbf{Input:} path, $\{\boldsymbol{p}_{x,\lambda}^{\mathrm{cons}}\}_{\lambda=1}^{n_\Lambda}$ where $x \in \{b, e\}$
\STATE \textbf{Output:} $\mathcal{S}$  \quad // $\mathcal{S}$ : Corridor
\STATE $\mathcal{S} \leftarrow \emptyset$, $\mathcal{Q} \leftarrow \emptyset$ \quad // $\mathcal{Q}$: Polyhedron queue
\FOR{$\lambda = 1$ \TO $n_\Lambda$}
    \STATE ($\mathcal{P}_1$, $\mathcal{P}_2$) $\leftarrow$ $\textsc{ActiveGenerate}$($\boldsymbol{p}_{x,\lambda}^{\mathrm{cons}}$)
    \STATE $\mathcal{Q}$.push($\mathcal{P}_1$), $\mathcal{Q}$.push($\mathcal{P}_2$)
\ENDFOR
\STATE $i \leftarrow 1$
\WHILE{$i < |\text{path}|$}
    \FOR{$j = i + 1$ \TO $|\text{path}|$}
        \IF{\textsc{IsFarthestPoint}(path[$i$], path[$j$])}
            \STATE $\mathcal{S}.\text{push}($\textsc{Generate}(path[$i$], path[$j$])$)$
            \STATE $i \leftarrow$ \textsc{LastPointInPoly}(path, $\mathcal{S}[-1]$), \textbf{break}
        \ENDIF
        \IF{$\mathcal{Q} \neq \emptyset$ \AND \textsc{InsidePoly}($\mathcal{Q}[0]$, path[$j$])}
            \STATE $\mathcal{S}.\text{push}$($\mathcal{Q}.\text{pop}$()), $\mathcal{S}.\text{push}$($\mathcal{Q}.\text{pop}$())
            \STATE $i \leftarrow$ \textsc{LastPointInPoly}(path, $\mathcal{S}[-1]$), \textbf{break}
        \ENDIF
    \ENDFOR
\ENDWHILE
\RETURN $\mathcal{S}$
\end{algorithmic}
\end{algorithm}

\section{Active SFC Generation}\label{sec: active sfc generation}
Given the constrained waypoint positions $\boldsymbol{p}^{\text{cons}}_{b,\lambda}$ and $\boldsymbol{p}^{\text{cons}}_{e,\lambda}$ $(\lambda \in \{1,2,\cdots, n_\Lambda\})$, our objective is to generate a safe and feasible trajectory for the aerial manipulator while satisfying all waypoint constraints. However, according to \cite{liu_planning_2017}, waypoints should intersect with non-empty regions formed by the intersection of two consecutive polyhedra. To achieve this objective, we propose a modified SFC generation strategy that combines multi-stage path searching with an active polyhedron generation method at constrained points, which not only enables waypoint constraint implementation but also enlarges the solution space. The complete algorithm is presented in \Cref{alg:sfc_generation}.

\begin{figure}
\centering
\includegraphics[width=1\linewidth]{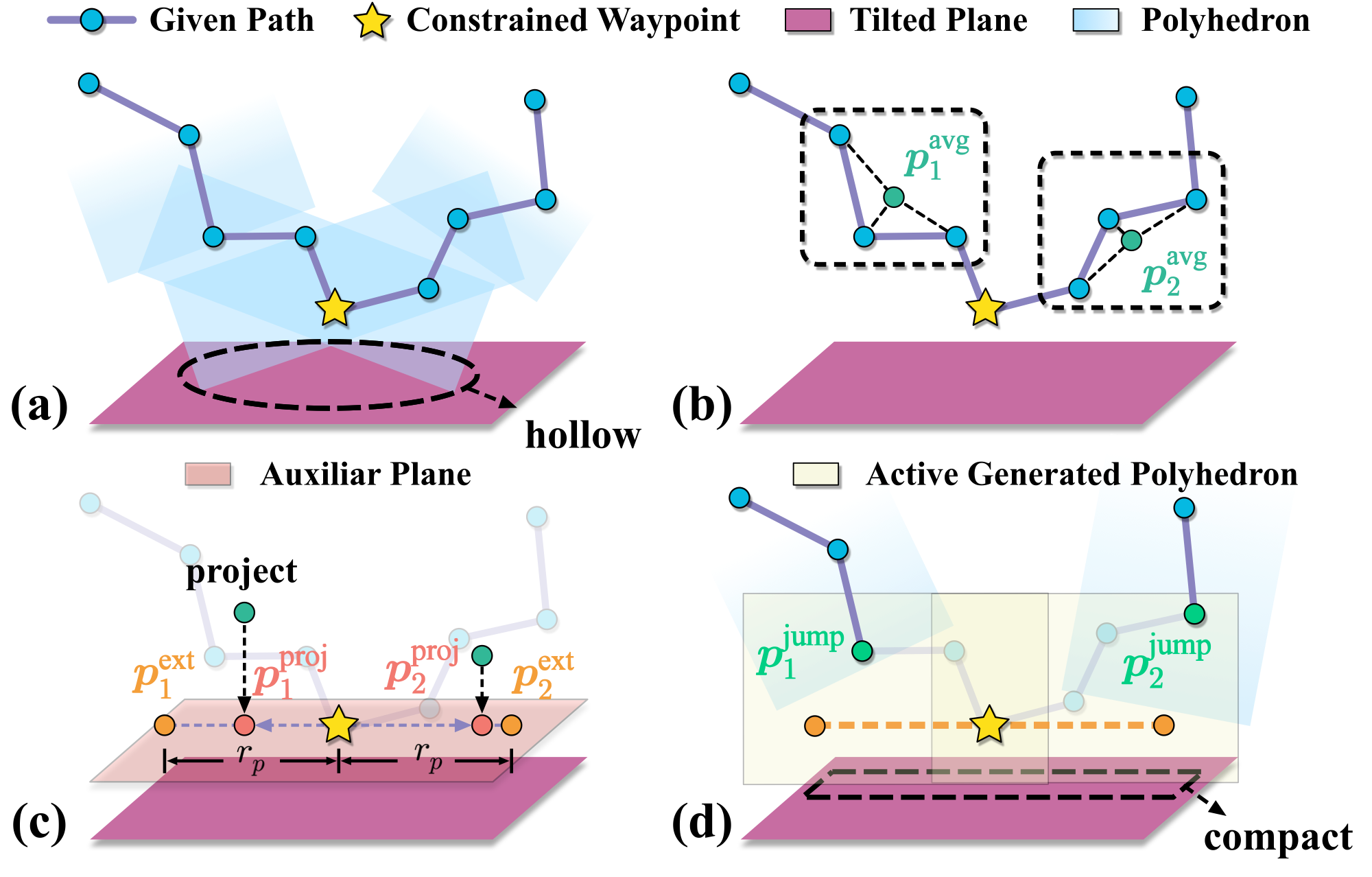}
\caption{Active polyhedrons generation algorithm: (a) hollow spaces from SFC generation on inclined surfaces; (b) computation of averaged points $\boldsymbol{p}_1^{\text{avg}}$ and $\boldsymbol{p}_2^{\text{avg}}$ from path points; (c) projection onto auxiliary planes yielding $\boldsymbol{p}_1^{\text{proj}}$, $\boldsymbol{p}_2^{\text{proj}}$ and extensions $\boldsymbol{p}_1^{\text{ext}}$, $\boldsymbol{p}_2^{\text{ext}}$; (d) polyhedron generation using constrained waypoints with points $\boldsymbol{p}_1^{\text{ext}}$ and $\boldsymbol{p}_2^{\text{ext}}$ for compact adherence to the surface.}
\label{fig: SFC}
\end{figure}

\subsection{Active Polyhedron Generation}\label{sec: active polyhedron generation}
Prior to the SFC construction, we leverage existing path searching algorithms JPS \cite{liu_search-based_2018} to obtain a reference path that sequentially traverses through all waypoints $\boldsymbol{p}^{\text{cons}}_{b,\lambda}$ and $\boldsymbol{p}^{\text{cons}}_{e,\lambda}$ from the start to the end.
Traditional construction methods \cite{liu_planning_2017,han_fast-racing_2021} directly generate a sequence of continuously intersecting polyhedrons by identifying and linking the farthest unblocked points along the reference path. 
However, obstacle occlusion and environmental factors introduce significant uncertainties during the polyhedron generation process. These uncertainties make it difficult to accurately locate intersection regions and can lead to reduced feasible solution spaces, such as hollow areas that may form between surfaces and polyhedrons, as illustrated in \Cref{fig: SFC} (a).
To address the challenge, we develop an active polyhedron generation strategy that extends the polyhedron around constrained waypoints of interest, inspired by the method proposed in \cite{ren_online_2023}.

As illustrated in \Cref{fig: SFC} (a), we define an arbitrary and winding purple path that orderly connects the start point $\boldsymbol{p}_{b,o}$, several constrained waypoints ($\boldsymbol{p}^\text{cons}_{b,\lambda}$ or $\boldsymbol{p}^\text{cons}_{e,\lambda}$), and the end point $\boldsymbol{p}_{b,f}$. 
For clarity, we focus on the situation at one of the constrained points.

According to \cite{liu_planning_2017}, the convex polyhedron generation is strongly related to the input line segment, obstacle distribution, and the bounding box configuration.
This relationship naturally leads us to consider a key geometric constraint: constructing two line segments that connect the constrained waypoint to its previous and subsequent points can guarantee that the two corresponding polyhedrons form a non-empty intersection region containing the constrained waypoint.
More crucially, if the expanding ellipsoid's $x$-axis is aligned parallel to the inclined plane, the bounding box can be optimally oriented to maximize coverage of the available space on the slope, thereby enlarging the feasible solution space for trajectory optimization.
To determine the directions towards and away from the constrained points, we calculate the average of a given number of points before and after the constrained point along the path, denoted as $\boldsymbol{p}^{\text{avg}}_1$ and $\boldsymbol{p}^{\text{avg}}_2$ respectively, as shown in \Cref{fig: SFC} (b). 
These points are then projected onto an auxiliary plane parallel to the previous plane but passing through the constrained waypoint, resulting in two projected vectors $\boldsymbol{p}_1^{\text{proj}}$ and $\boldsymbol{p}_2^{\text{proj}}$ as illustrated in \Cref{fig: SFC} (c). 
After normalizing and scaling these vectors by $r_p$, we obtain two new generation points, $\boldsymbol{p}_1^{\text{ext}}$ and $\boldsymbol{p}_2^{\text{ext}}$. 
By connecting the constrained point to these generation points, we can directly generate two polyhedrons that are tightly fitted to the inclined plane, as illustrated in \Cref{fig: SFC} (d). These constrained polyhedrons are then stored in a temporary queue $\mathcal{Q}$.
Additionally, we define two jump points, $\boldsymbol{p}_1^{\text{jump}}$ and $\boldsymbol{p}_2^{\text{jump}}$, which are the first and last points along the path within these two polyhedrons respectively. Their specific applications will be discussed in the following subsection.

\subsection{SFC Generation}\label{sec: sfc generation}
After generating two polyhedrons for each constrained point as described above, the complete SFC is constructed through an iterative process along the path. 
During this iteration, when the current point reaches a predefined threshold distance from the last point, the algorithm pauses to generate a new polyhedron and then resumes iteration from the last point within this newly generated polyhedron.
When the iterative point reaches $\boldsymbol{p}_1^{\text{jump}}$, the iteration immediately pauses, and the two constrained polyhedrons are popped from the temporary queue $\mathcal{Q}$ and incorporated into the SFC. The iteration then resumes from the corresponding $\boldsymbol{p}_2^{\text{jump}}$ and continues the standard generation process.
Ultimately, we obtain the SFC in $\mathcal{H}$-representation as a series of consecutively intersected polyhedrons \cite{yang_whole-body_2021}:
\begin{subequations}
\begin{align}
&\mathcal{S}:=\bigcup_{i=1}^{M} \mathcal{H}_i \subset \mathbb{R}^3 \label{eq: SFC}\\
&\mathcal{H}_i:=\left\{\boldsymbol{x} \in \mathbb{R}^3 \mid \mathbf{A}_{\boldsymbol{i}} \boldsymbol{x} \preceq \boldsymbol{b}_{\boldsymbol{i}}\right\}
\end{align}
\end{subequations}
where $M$ is the total number of polyhedrons in the SFC.

\section{Basic Problem Formulation}\label{sec: Problem Formulation}
This section presents our basic optimization problem formulation for aerial manipulators, which guarantees trajectory smoothness, safety, and dynamic and kinematic feasibility without considering task-specific waypoint constraints.
We begin by formulating a comprehensive objective function that balances control effort and time consumption metrics. The objective function is subject to multiple constraint categories: collision-free constraints implemented through a varying ellipsoid, and maximum velocity, acceleration, thrust, body rate, and workspace constraints.

\subsection{Objective Function}
We focus on generating a 6-dimensional trajectory $\boldsymbol{p}(t) = [\boldsymbol{p}_b^\top(t), {}^{\mathcal{D}}\boldsymbol{p}_e^\top(t)]^\top\in\mathbb{R}^6$ for an aerial manipulator.
As discussed in \Cref{sec: trajectory_representation}, we decompose the complete trajectory into intermediate points and intervals. For convenience, we separate these intermediate points for the quadrotor and the manipulator into:
\begin{align}
\mathbf{P} &= \left[ \boldsymbol{p}_{b,0},\boldsymbol{p}_{b,1}, \cdots, \boldsymbol{p}_{b,M} \right]^\top \in \mathbb{R}^{(M+1)\times 3},\\
\mathbf{Q} &= \left[{}^\mathcal{D}\boldsymbol{p}_{e,0},{}^\mathcal{D}\boldsymbol{p}_{e,1}, \cdots, {}^\mathcal{D}\boldsymbol{p}_{e,M} \right]^\top \in \mathbb{R}^{(M+1)\times 3}.
\end{align}  
Accordingly, the coefficient matrix is also decomposed as $\mathbf{C} = [\mathbf{c}_b, \mathbf{c}_e]$.
Combined with the previously mentioned time vector $\mathbf{T}$, we can formulate the optimization problem as:
\begin{subequations}
\begin{align}
\min_{\mathbf{P},\mathbf{Q}, \mathbf{T}} &\; \mathcal{J} =
\int_0^{T_\sigma} \| \boldsymbol{p}^{(s)}(t)\|^2 \mathrm{d}t + \rho T_\sigma \label{eq:obj} \\
\text{s.t.} \quad &\mathcal{G} (\boldsymbol{p}(t),\ldots,\boldsymbol{p}^{(s)}(t)) \preceq \mathbf{0},\; \forall t \in[0, T_\sigma] \label{eq:constraint1} \\
& \boldsymbol{p}_1^{[s-1]}(0) = \mathbf{p}_o, \quad  \boldsymbol{p}_M^{[s-1]}(T_M) =\mathbf{p}_f \label{eq:constraint2} \\
&T_i > 0,\quad i\in\{1,2,\cdots,M\} \label{eq:constraint4}
\end{align}
\end{subequations}
where $T_\sigma = \sum_{i=1}^{M}T_i$ is the total time consumption with regularization factor $\rho>0$, and $\mathcal{G}$ denotes the inequality constraints over $[0,T_\sigma]$.
The notation $\boldsymbol{p}^{[s-1]}$ represents the multi-order differentials matrix $[\boldsymbol{p}, \dot{\boldsymbol{p}}, \ldots, \boldsymbol{p}^{(s-1)}] \in \mathbb{R}^{6\times s}$ . 
Let $\mathbf{p}_o, \mathbf{p}_f \in \mathbb{R}^{6 \times s}$ denote the initial and terminal conditions for the aerial manipulator in $\mathcal{F}_\mathcal{W}$ and the end-effector in $\mathcal{F}_\mathcal{D}$.
$T_i$ denotes the duration of the $i$-th segment, and $M$ is the total number of segments. 

The inequality constraints in \Cref{eq:constraint1} encompass multiple aspects, including safety, dynamic, kinematic and attitude constraints. We adopt a penalty-based approach to handle these constraints efficiently by incorporating them into the objective function with appropriate weights, thereby transforming the original constrained optimization problem into an unconstrained one.
To reduce the computational complexity, we approximate the continuous integral using discrete numerical integration:
\begin{equation}
\int_0^{T_\sigma} J_\mathcal{G} \mathrm{d} t \approx \sum_{i=1}^M \sum_{n=0}^{N-1}\frac{T_i}{N} J_\mathcal{G}^i(\frac{nT_i}{N}) \label{eq:discrete}
\end{equation}
where $J_\mathcal{G}^i$ represents the soft penalty terms for the $i$-th segment, and $N$ denotes the number of discretization points per segment.

\subsection{Control Effort and Total Time Penalty}
We define the control effort penalty $J_c$ as the integral of the squared third-order derivatives (i.e., minimum jerk) for both the aerial manipulator and the end-effector. The total control effort is given by:
\begin{equation}
J_c = \int_{0}^{T_\sigma}  \left\|\boldsymbol{p}^{(3)}(t)\right\|^2 
 \mathrm{d}t = \sum_{i=1}^{M} \int_0^{T_i} \left\|\boldsymbol{p}_{i}^{(3)}(t)\right\|^2 \mathrm{d}t .
\end{equation}
% \textcolor{red}{\st{The partial derivatives of the $i$-th trajectory segment with respect to $\mathbf{c}_{x,i}$ and $T_i$ can be derived as follows (where $x\in\{b,e\}$):}}
% \textcolor{red}{
% \begin{align}
% \cancel{\frac{\partial J_c^i}{\partial \mathbf{c}_{x,i}} = 2 \int_{0}^{T_i} \boldsymbol{\beta}^{(3)}(t) \boldsymbol{\beta}^{(3)}(t)^\top \mathbf{c}_{x,i} \, \mathrm{d}t}, \\
% \cancel{\frac{\partial J_c^i}{\partial T_i} =\sum_{x} \left( \mathbf{c}_{x,i}^\top
% \boldsymbol{\beta}^{(3)}(T_i) \right)^\top  \mathbf{c}_{x,i}^\top \boldsymbol{\beta}^{(3)}(T_i)}.
% \end{align}
% }

In addition, to ensure trajectory efficiency, we incorporate a time penalty term $J_t$, defined as the total trajectory duration:
\begin{equation}
J_t =T_\sigma =  \sum_{i=1}^M T_i,
\end{equation}
where $T_i$ represents the duration of the $i$-th trajectory segment.

% \begin{figure}[t]
%     \centering
%     \includegraphics[width=1\linewidth]{assets/ellipsoid.png}
%     \caption{Dynamic ellipsoid approximation results for different shapes, where the horizontal semi-major axis length $r_e$ remains constant, while the vertical axis length $h(t)$ varies dynamically.}
%     \label{fig:ellipsoid}
% \end{figure}

\begin{figure}
    \centering
    \includegraphics[width=1\linewidth]{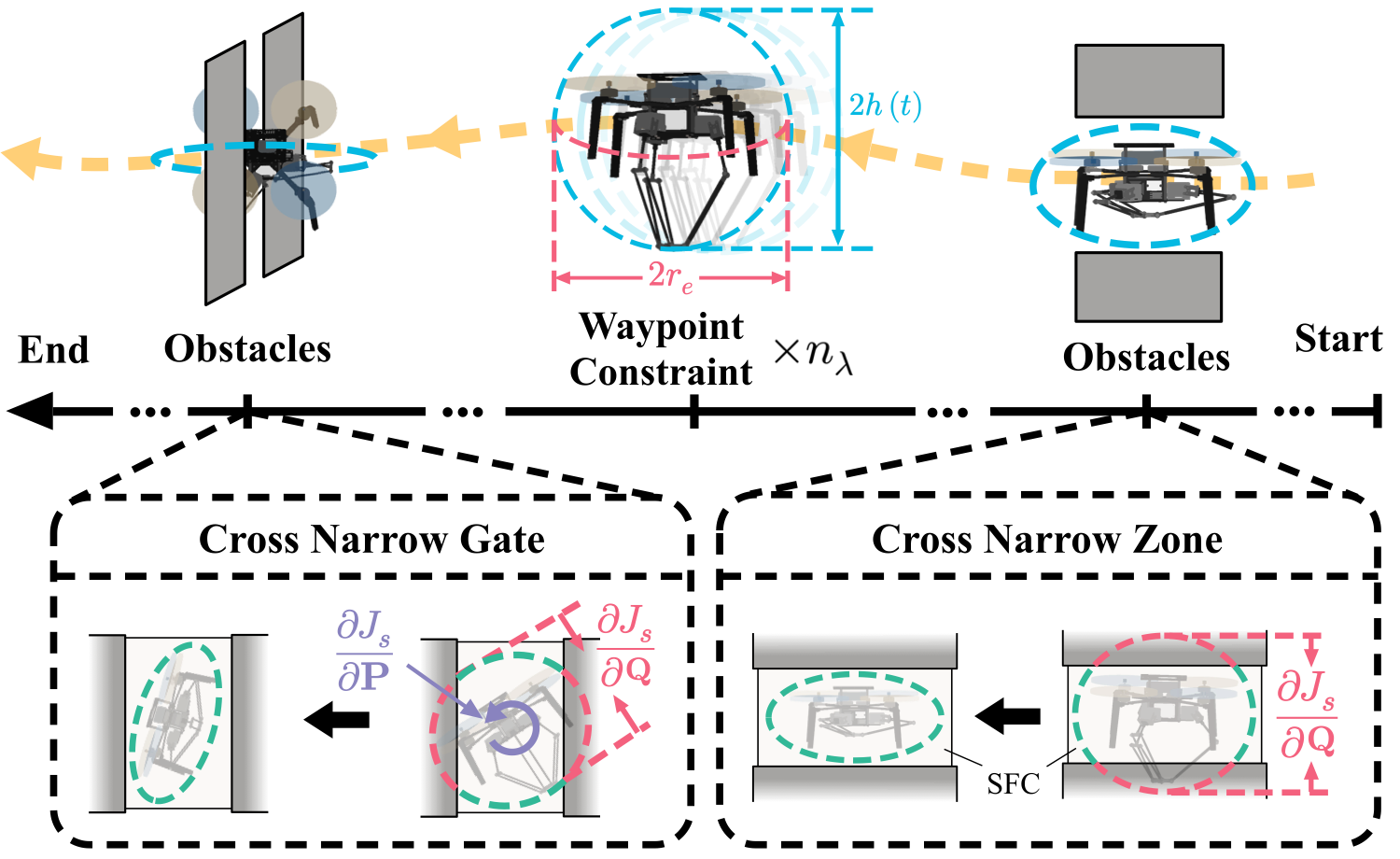}
    \caption{Visualization of the safety penalty mechanism: The aerial manipulator achieves whole-body collision avoidance by optimizing its configuration through orientation adjustment and delta arm extension, while maintaining task constraints at specified waypoints within the SFC limits.}
    \label{fig:safety}
\end{figure}

\subsection{Safety Penalty}\label{safety}
We extend the approaches proposed by \cite{yang_whole-body_2021} and \cite{liu_planning_2017} by modeling the entire aerial manipulator system as a varying ellipsoid during task execution. 
The ellipsoid, designed with fixed semi-major axes of length $r_e$ in the $xy$-plane, incorporates a variational function $h(t)$ to dynamically adjust its length along the $z$-axis, accounting for the manipulator's motion range and ensuring complete coverage of the entire system, as shown in \Cref{fig:safety}.
The formulation of this ellipsoidal model is given by:
\begin{subequations}
\begin{align}
\mathcal{E}(\boldsymbol{\xi}, t) &= \mathbf{G}(t) \cdot (\mathbf{R}_{\mathcal{B}}(t)\boldsymbol{\xi}) + \boldsymbol{p}_b(t), \quad \boldsymbol{\xi}^\top \boldsymbol{\xi} \le 1, \\ \label{G(t)}
\mathbf{G}(t) &= \mathrm{diag}(r_e, r_e, h(t)) \in \mathbb{R}_{> 0}^{3 \times 3}, \\ \label{eq: ht}
h(t) &= \left( {}^{\mathcal{D}}\boldsymbol{p}_{\mathcal{B}} - {}^{\mathcal{D}}\boldsymbol{p}_{e}(t) \right)^\top \boldsymbol{e}_3,
\end{align}
\end{subequations}
where $\boldsymbol{\xi}\in \mathbb{R}^3$ represents any vector within a unit sphere, and $\mathcal{E}$ represents the approximated ellipsoid in the world frame $\mathcal{F}_\mathcal{W}$. The matrix $\mathbf{G}(t)$ encodes the semi-lengths of the ellipsoid's three axes along its diagonal.
% $\mathbf{R}_{\mathcal{B}}(t) \in SO(3)$ and $\boldsymbol{p}_b(t) \in \mathbb{R}^3$ represent the rotational and translational components of the ellipsoid's configuration, respectively, 
$\boldsymbol{e}_3 = [0,0,1]^\top$ is the unit vector. 
${}^{\mathcal{D}}\boldsymbol{p}_{\mathcal{B}}$ denotes the translation from $\mathcal{F}_\mathcal{D}$ to $\mathcal{F}_\mathcal{B}$.

To guarantee safety, we constrain the dynamic ellipsoid to remain within the SFC using the aforementioned methodologies, as shown in \Cref{fig:safety}. This containment constraint can be mathematically expressed as:
\begin{equation}
\mathcal{E} \subset \mathcal{S}
\end{equation}
where $\mathcal{S}$ is the space defined in \Cref{eq: SFC}.
To achieve this, we define the following penalty functions:
\begin{subequations}
\begin{align}
J_s &=  \sum_{i=1}^{M} 
\sum_{n=0}^{N-1}
\sum_{k=1}^{K_i} 
\frac{T_i}{N} 
\mathcal{K}  \left(  \mathcal{P}_{s, i, n, k}\right),  \label{eq: J_s} \\ 
\mathcal{P}_{s, i, n, k} &=  \left(\boldsymbol{p}_{i,n}^k(\tau) - \boldsymbol{\hat{p}}_i^k\right)^\top
\boldsymbol{\hat{n}}_i^k + d_s \label{eq: P_sink}
\end{align}
\end{subequations}
where $\tau= \frac{nT_i}{N}$, a definition that holds for all subsequent derivations. Here, $\mathcal{K}(\cdot)$ denotes the cubic smooth function defined as $\mathcal{K}(x) = \max(x^3, 0)$. For the $i$-th polyhedron containing $K_i$ half-planes, $\boldsymbol{\hat{p}}_i^k$ represents an arbitrary point on the $k$-th half-plane with $\boldsymbol{\hat{n}}_i^k$ as its outward-pointing normal vector, and $d_s$ defines a small safety margin. The term $\boldsymbol{p}_{i,n}^k$ represents the point on the ellipsoid's surface where its normal vector is perpendicular to the $k$-th plane. This point can be solved by inverting the ellipsoidal tangent point equation and be computed as follows:
\begin{equation}\label{eq: tangent}
\left\{
\begin{aligned}
 &\boldsymbol{p}_{i,n}^k(\tau) = \mathbf{R}_{\mathcal{B},i}(\tau) \cdot {}^{\mathcal{B}}\boldsymbol{p}_{i,n}^k(\tau) + \boldsymbol{p}_{b,i}(\tau), \\[0.5em]
 &{}^{\mathcal{B}}\boldsymbol{p}_{i,n}^k(\tau) = \pm \frac{\mathbf{G}^2(\tau) \cdot (\mathbf{R}_{\mathcal{B},i}^{\top} (\tau) \boldsymbol{\hat{n}}_{i}^k)}{\left\| \mathbf{G}(\tau)  \cdot (\mathbf{R}_{\mathcal{B},i}^{\top} (\tau) \boldsymbol{\hat{n}}_i^k) \right\|_2}, \\[0.5em]
 &{}^{\mathcal{B}}\boldsymbol{p}_{i,n}^k(\tau) \cdot \left(\mathbf{R}_{\mathcal{B},i}^{\top}(\tau) \cdot
 \boldsymbol{\hat{n}}_i^k\right) \ge 0.
\end{aligned}
\right.
\end{equation}
Then, we can easily derive the gradient of the safety cost with respect to the optimization variables $\mathbf{P}$, $\mathbf{Q}$, and $\mathbf{T}$.

\subsection{Kinematics and Dynamics Penalty}\label{sec: dyna}
To ensure the physical feasibility of the generated trajectories, we need to consider both kinematic and dynamic constraints. The kinematic constraints govern the system's workspace, velocities, and body rates, while the dynamic constraints regulate the thrust forces.

We first consider the kinematic constraints. For the delta arm to operate safely and effectively, its end-effector must remain within a prescribed workspace. Following \cite{cao_motion_2024}, we model this workspace as a cuboid in the Cartesian coordinate system, where $\kappa \in\{x,y,z\}$ denotes any of the three axes. The workspace is bounded by:
\begin{subequations}
\begin{align}
\kappa_{e,\ell} \leq \boldsymbol{e}_\kappa^\top {}^{\mathcal{D}}\boldsymbol{p}_e(\tau) \leq \kappa_{e,u},
\end{align}
\end{subequations}
where $\kappa_{e,\ell}$ and $\kappa_{e,u}$ denote the lower and upper bounds along axis $\kappa$, respectively, and $\mathbf{e}_\kappa$ represents the unit vector along the $\kappa$-axis. To enforce these workspace constraints, we introduce the following penalty terms:
\begin{subequations}
\begin{align}
&J_{k} = \sum_{i=1}^{M}\sum_{n=0}^{N-1}\sum_{\kappa} \frac{T_i}{N}\Big( \mathcal{K} \left(
\mathcal{P}_{k_\ell,i,n,\kappa}\right) + \mathcal{K}\left( \mathcal{P}_{k_u,i,n,\kappa} \right) \Big),\\
&\mathcal{P}_{k_\ell,i,n,\kappa} = \left( \boldsymbol{e}_\kappa^\top{}^{\mathcal{D}}\boldsymbol{p}_{e,i}(\tau)  \right)^2 - \kappa_{e,u}^2, \\
&\mathcal{P}_{k_u,i,n,\kappa} = \kappa_{e,\ell}^2 - \left( \boldsymbol{e}_\kappa^\top {}^{\mathcal{D}}\boldsymbol{p}_{e,i}(\tau)  \right)^2.
\end{align}
\end{subequations}

To prevent excessive motion, we also constrain the velocities of both the aerial manipulator and the end-effector. The aerial manipulator's velocity is limited to an upper bound $v_{b,u}$, while the end-effector velocity is constrained by $v_{e,u}$. These constraints are incorporated through the following penalty terms:
\begin{subequations}
\begin{align}
   J_{k,vel} &= \sum_{i=1}^{M}\sum_{n=0}^{N-1} \frac{T_i}{N}
   \Big(\mathcal{K} \left( \mathcal{P}_{b,v,i,n}\right)+
   \mathcal{K} \left( \mathcal{P}_{e,v,i,n}\right)
   \Big),\\
   \mathcal{P}_{b,v,i,n} &= \left\| \dot{\boldsymbol{p}}_{b,i}(\tau) \right\|^2 - v_{b,u}^2,\\
   \mathcal{P}_{e,v,i,n} &= \left\| ^{\mathcal{D}}\dot{\boldsymbol{p}}_{e,i}(\tau) \right\|^2 - v_{e,u}^2,
\end{align}
\end{subequations}
where $\dot{\boldsymbol{p}}_{b,i}(\tau)$ represents the velocity of the aerial manipulator in $\mathcal{F}_\mathcal{W}$, and $^{\mathcal{D}}\dot{\boldsymbol{p}}_{e,i}(\tau)$ represents the velocity of the end-effector in $\mathcal{F}_\mathcal{D}$.

As another kinematic constraint, we consider the aerial manipulator's body rates. Given that yaw angle variations are neglected in our framework, we focus only on the body rates in the $xy$-plane. The corresponding penalty terms are defined as:
\begin{subequations}
\begin{align}
J_{k, bdr} & =\sum_{i=1}^M \sum_{n=0}^{N-1} \frac{T_i}{N}\mathcal{K}\left(\mathcal{P}_{\omega, i, n}\right), \\
\mathcal{P}_{\omega, i, n} & =\left\|{}^{\mathcal{B}}\boldsymbol{\omega}_{xy,i}\right\|^2-\omega_{xy,u}^2,
\end{align}
\end{subequations}
where ${}^{\mathcal{B}}\boldsymbol{\omega}_{xy,i} =\left[\omega_x, \omega_y\right]^\top$ represents the body angular rates in the $xy$-plane, and $\omega_{xy,u}$ denotes the upper limit of the body rates. 

For dynamic constraints, we consider the aerial manipulator's thrust capabilities. The thrust $f$ must remain within physically realizable bounds:
\begin{equation}
   f_\ell\le f \le f_u,
\end{equation}
where $f_\ell$ and $f_u$ denote the lower and upper thrust limits. We formulate the corresponding penalty terms as:
\begin{subequations}
\begin{align}
J_{d,thr} &=  \sum_{i=1}^{M}\sum_{n=0}^{N-1}\frac{T_i}{N} \Big(\mathcal{K} \left( 
\mathcal{P}_{f_u,i,n} \right)+ \mathcal{K} \left( 
\mathcal{P}_{f_\ell,i,n}
\right) \Big),\\
\mathcal{P}_{f_u,i,n} &= \left\| m_c(\ddot{\boldsymbol{p}}_{b,i}(\tau) + g\boldsymbol{z}_\mathcal{W}) -
\boldsymbol{f}_{\mathrm{ext}}
\right\|^2 - f_u^2, \\
\mathcal{P}_{f_\ell,i,n} &= f_\ell^2 - \left\| m_c(\ddot{\boldsymbol{p}}_{b,i}(\tau) + g\boldsymbol{z}_\mathcal{W}) -
\boldsymbol{f}_{\mathrm{ext}}
\right\|^2,
\end{align}
\end{subequations}
where $\ddot{\boldsymbol{p}}_{b,i}(\tau)$ denotes the aerial manipulator's acceleration.

The gradients of all kinematics and dynamics penalty terms with respect to the optimization variables $\mathbf{P}$, $\mathbf{Q}$ and $\mathbf{T}$ can be efficiently computed using the method described in \cite{han_fast-racing_2021, gao_adaptive_2024}.

With the complete formulation of constraints and their corresponding gradients for the aerial manipulator's basic flight planning, we can solve the optimization problem using L-BFGS to obtain collision-free flight trajectories for the aerial manipulator.

\section{Waypoint Constraints and IL-Guided Optimization}\label{sec: Waypoint_Constraint}
\begin{figure}
    \centering
    \includegraphics[width=1.0\linewidth]{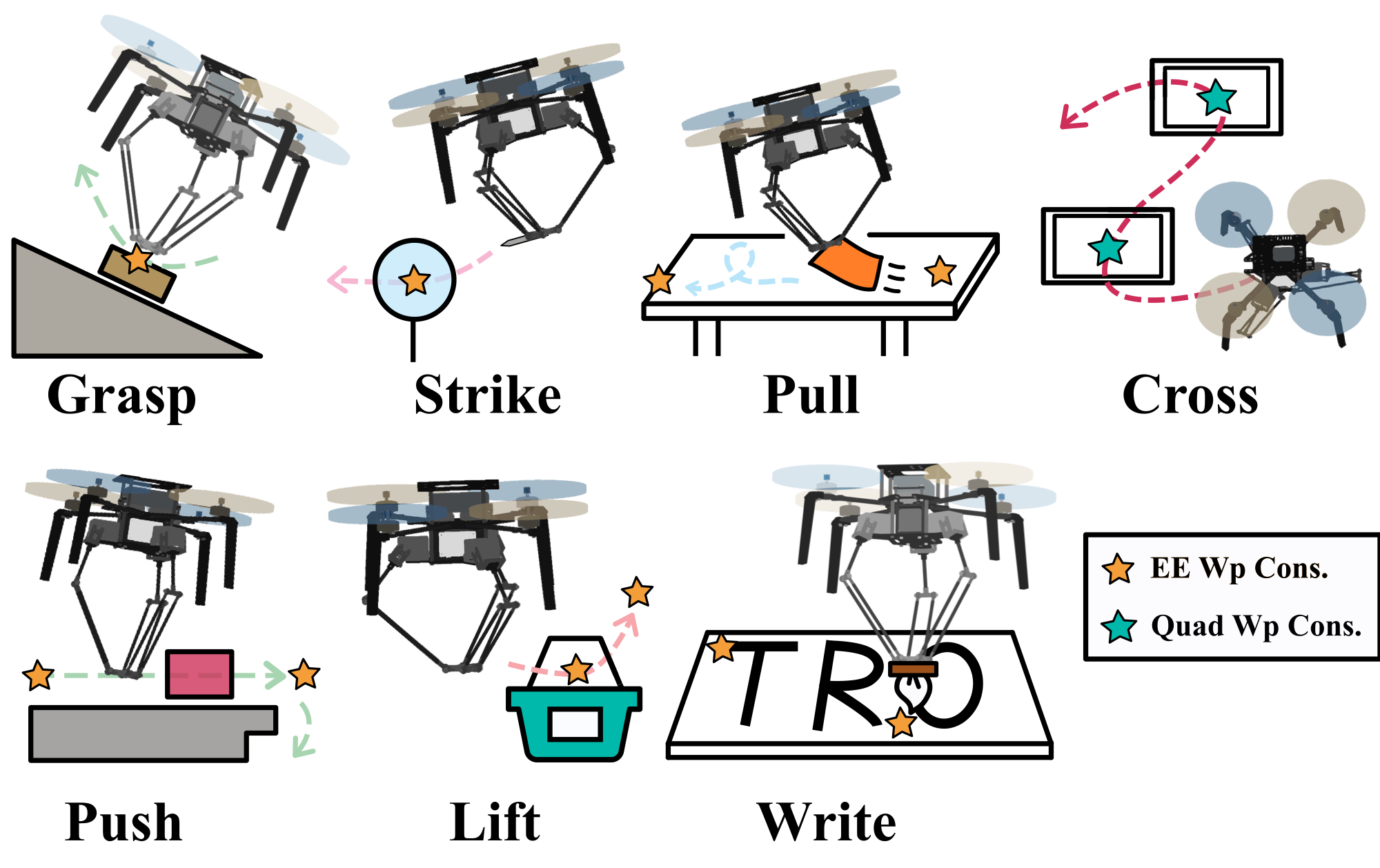}
    \caption{Illustration of representative manipulation primitives achieved through waypoint constraints. The figure demonstrates various fundamental skills including \textit{Grasp}, \textit{Strike}, \textit{Pull}, \textit{Cross}, \textit{Push}, \textit{Lift}, and \textit{Write}. All these manipulation tasks can be effectively accomplished by implementing appropriate waypoint constraints combined with motion restrictions during execution.}
    \label{fig:subskill}
\end{figure}

Having established the basic optimization formulation in the previous section, we now turn to the incorporation of task-specific waypoint constraints, which can be applied to either the quadrotor or the end-effector in the frame $\mathcal{F}_\mathcal{W}$ independently, as shown in \Cref{fig:subskill}. These constraints include higher-order velocity and orientation requirements at designated waypoints. Building upon this framework, we further propose an IL-guided optimization approach to improve performance for specific manipulation tasks.

\subsection{Waypoint Constraints Implementation}
Based on our modified SFC generation strategy, each constrained waypoint corresponds to an intersection of two polyhedrons and an intermediate point. We define the mapping
\begin{equation}
\phi: \{1, \ldots, n_\Lambda\} \rightarrow \{1, \ldots, M-1\}
\end{equation}
to map constrained waypoint indices to their corresponding intermediate point indices.

\textbf{1) Quadrotor Waypoint Constraints}: 
Since the quadrotor trajectory is planned in frame $\mathcal{F}_\mathcal{W}$, we exploit the spatio-temporal trajectory decomposition property by fixing certain waypoints as constants while maintaining others as optimization variables. The quadrotor waypoint constraints are denoted as:
\begin{equation}
\mathbf{P}_{b}^{\mathrm{cons}} := \{ 
\boldsymbol{p}_{b, \phi(\lambda) } = 
\boldsymbol{p}^{\mathrm{cons}}_{b, \lambda} 
\mid \lambda \in \Lambda_b\}.
\end{equation}
The modified optimization variables become:
\begin{equation}
\bar{\mathbf{P}} = \mathbf{P} \setminus \mathbf{P}_{b}^{\mathrm{cons}}
\end{equation}
where $\setminus$ denotes set difference. Due to the spatio-temporal trajectory properties, the resulting trajectory is guaranteed to traverse all constrained waypoints in $\mathbf{P}_{b}^{\mathrm{cons}}$.

\textbf{2) End-Effector Waypoint Constraints}: 
The end-effector trajectory is planned in $\mathcal{F}_\mathcal{D}$ and cannot be directly fixed like the quadrotor trajectory. Instead, we enforce end-effector waypoint constraints through a soft penalty approach. We have generated polyhedron intersections near each $\boldsymbol{p}_{e, \lambda}^{\mathrm{cons}}$ to ensure the corresponding trajectory intermediate points pass through their vicinity. Therefore, using the end-effector's world frame position from \Cref{eq: ee_world_representation}, we formulate the penalty term:
\begin{align}
J_e &= \sum_{\lambda\in \Lambda_e} \left\|\boldsymbol{p}_{e,\phi(\lambda)} - \boldsymbol{p}_{e, \lambda}^{\mathrm{cons}}\right\|^2.
\end{align}

\subsection{Task-Specific Constraints}
\textbf{1) Axis-Wise End-Effector Motion Constraint}:
When executing specific manipulation skills such as pulling, pushing, and writing tasks, the aerial manipulator's end-effector must maintain its position parallel to a defined plane \cite{dimming_non_prehensile_2024}. More restrictively, operations involving sliding mechanisms require the end-effector to remain constrained to a line throughout the motion. These geometric constraints pose significant challenges for trajectory planning, as conventional point-to-point planning methods cannot guarantee such continuous spatial restrictions.

To address these requirements, we introduce axis-wise end-effector position motion constraints that enforce planar or linear motion during designated trajectory segments. The constraints are formulated as follows:
\begin{equation}
J_{pe} = \sum_{i=1}^M  \mathcal{I}_p(i) \sum_{n=0}^{N-1}  \frac{T_i}{N} \|   \boldsymbol{c}_p^\top(\boldsymbol{p}_{e,i}(\tau) - \boldsymbol{p}_{\mathrm{des},\lambda}) \|^2
\end{equation}
where $\mathcal{I}_p(i)\in \{0, 1\}$ that activates constraints for specific trajectory segments, and $\boldsymbol{c}_p \in \mathbb{R}^3$ is a user-defined binary vector where setting any dimension to 1 indicates that the corresponding coordinate should be constrained. These axis-wise motion constraints are typically employed in conjunction with two end-effector waypoint constraints to ensure that the constrained dimensions maintain identical values between consecutive waypoints.

\textbf{2) Velocity Constraint}:
In certain manipulation tasks, the end-effector's velocity must be constrained to ensure precise operation. For instance, during object grasping maneuvers, the end-effector may need to maintain zero velocity or a small downward velocity along the $\boldsymbol{z}_\mathcal{B}$ to achieve stable contact.

The velocity constraint is formulated as:
\begin{equation}
J_{ve} =  \sum_{\lambda\in\Lambda_e}   \mathcal{I}_v(\lambda) \|  \boldsymbol{c}_v^\top(\dot{\boldsymbol{p}}_{e,\phi(\lambda)} - \dot{\boldsymbol{p}}_{\mathrm{des},\lambda}) \|^2
\end{equation}
where $\mathcal{I}_v(\lambda) \in \{0,1\}$ is a binary selector that activates velocity constraints for waypoint $\lambda$, $\boldsymbol{c}_v\in \mathbb{R}^3$ is a user-defined binary vector specifying which velocity components to constrain, and $\dot{\boldsymbol{p}}_{\mathrm{des},\lambda}$ is the desired end-effector velocity at the constrained waypoint.

\textbf{3) Orientation Constraint}:
Since the end-effector in the delta arm configuration shares the same orientation as the quadrotor, orientation constraints can be naturally imposed on the quadrotor's intermediate waypoints. This inherent coupling eliminates the need to distinguish between quadrotor and end-effector orientations. The orientation constraint is formulated as:
% \begin{equation}
% J_{oe} =  \sum_{\lambda\in\Lambda_e}   \mathcal{I}_o(\lambda) \| (m_c(\ddot{\boldsymbol{p}}_{e,\phi(\lambda)} + g\boldsymbol{z}_\mathcal{W}) - \boldsymbol{f}_\mathrm{ext} - f\boldsymbol{o}_{\mathrm{des},\lambda}) \|^2
% \end{equation}
% where $\mathcal{I}_o(\lambda) \in \{0,1\}$ is a binary selector for activating orientation constraints, $f$ is the thrust and $\boldsymbol{o}_{\mathrm{des},\lambda}$ is the desired orientation at $\lambda$-th waypoint. 
\begin{equation}
J_{oe} = \sum_{\lambda\in\Lambda_e} \mathcal{I}_o(\lambda) \| \boldsymbol{f}_{\lambda} - \|\boldsymbol{f}_{\lambda}\|\boldsymbol{o}_{\mathrm{des},\lambda} \|^2
\end{equation}
where $\mathcal{I}_o(\lambda) \in \{0,1\}$ is a binary selector for activating orientation constraints, $\boldsymbol{f}_{\lambda} = m_c(\ddot{\boldsymbol{p}}_{b,\phi(\lambda)} + g\boldsymbol{z}_\mathcal{W}) - \boldsymbol{f}_\mathrm{ext}$ is the actual thrust vector,   and $\boldsymbol{o}_{\mathrm{des},\lambda}$ is the desired unit orientation vector at the $\lambda$-th waypoint.

% 放一些任务的图，然后指出这些任务是可以通过设置不同类型的waypoint constraint。

\begin{figure}[t]
    \centering
    \includegraphics[width=1\linewidth]{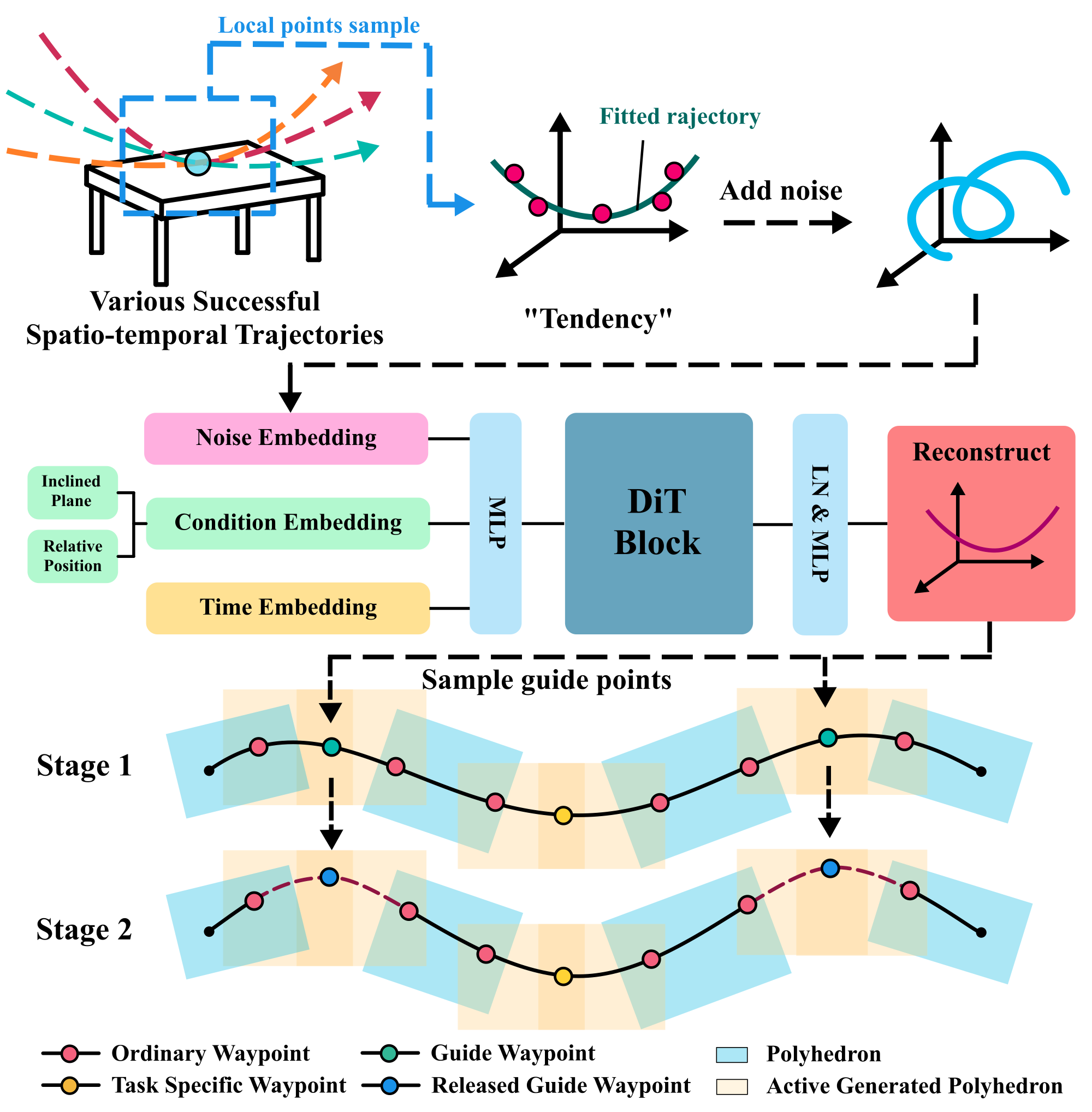}
    \caption{Visualization of the IL-guided optimization framework. The process begins with collecting various successful spatio-temporal trajectories from demonstration data. These trajectories are fitted using polynomial approximation to extract motion tendencies. A DiT model learns to reconstruct these motion tendencies through noise embedding, condition embedding, and time embedding. The optimization process consists of two stages: Stage 1 employs guide waypoints as hard constraints; Stage 2 releases these hard constraints, allowing free optimization.}
    \label{fig:waypointconstraint_new}
\end{figure}

\begin{algorithm}[tbp]
\caption{IL-guided Optimization}
\label{alg:IL_guided_optimization}
\begin{algorithmic}[1]
\STATE \textbf{Input:} $\boldsymbol{p}_{o}$, $\boldsymbol{p}_{f}$, $\boldsymbol{p}_{\text{inc}}$, 
$\boldsymbol{o}_{\text{inc}}$,  $\epsilon_{\text{loose}}$, $\epsilon_{\text{strict}}$
\STATE \textbf{Output:} $\boldsymbol{p}(t), t\in[0, T_\sigma]$
\STATE $\boldsymbol{a}$  $\leftarrow$ \textsc{ImitationLearning}(
$\boldsymbol{p}_{\text{inc}}$, $\boldsymbol{o}_{\text{inc}}$, $\boldsymbol{p}_{o}$)
\STATE $\boldsymbol{w}_{\text{guide}}$ $\leftarrow$ \textsc{SamplePoints}($\boldsymbol{a}$) 
\STATE $\boldsymbol{w}_{\text{task}}$ $\leftarrow$ \textsc{TaskSpecific}($\boldsymbol{p}_{\text{inc}}$, $\boldsymbol{o}_{\text{inc}}$)
\STATE $\boldsymbol{x}_1$ $\leftarrow$ \textsc{Init}($\boldsymbol{p}_{o}$, [$\boldsymbol{w}_{\text{guide}}$, $\boldsymbol{w}_{\text{task}}$], $\boldsymbol{p}_{f}$)
\STATE $\boldsymbol{x}_2$ $\leftarrow$ \textsc{Optimization}($\boldsymbol{x}_1$, $\epsilon_{\text{	loose}}$, [$\boldsymbol{w}_{\text{guide}}$, $\boldsymbol{w}_{\text{task}}$])
\STATE // Release the IL guided constraints
\STATE $\boldsymbol{x}_3$ $\leftarrow$ \textsc{Optimization}($\boldsymbol{x}_2$, $\epsilon_{\text{strict}}$, $\boldsymbol{w}_{\text{task}}$)
\RETURN \textsc{Polynomial}($\boldsymbol{x}_3$)
\end{algorithmic}
\end{algorithm}

\subsection{IL Prior}

Several complex manipulation tasks requiring extreme orientations, such as inclined surface grasping and striking, present significant challenges for trajectory optimization. These tasks involve inherently conflicting constraints - for example, orientation requirements conflict with maximum body rate limitations, and zero end-effector velocity during grasping conflicts with time penalty objectives. The complexity of these constraint interactions significantly amplifies the non-convex nature of the optimization problem. Since optimization fundamentally involves balancing competing objectives, poor initial solutions can lead the optimizer to converge to local optima that completely sacrifice task constraints \cite{ni_robust_2021}.

We focus on two tasks requiring aggressive maneuvers: inclined grasping and inclined striking.
For these complex tasks, designing simple waypoint constraints $\boldsymbol{w}_\text{task}$ that contain position $\boldsymbol{p}_\text{inc}$ and corresponding orientations $\boldsymbol{o}_\text{inc}$ often results in poor local optima that fail to satisfy overall task requirements.
To address this challenge, we analyze successful trajectories from both tasks and discover that they exhibit distinct flight "tendencies."
For example, striking tasks consistently demonstrate arc-like trajectory patterns.
These observed patterns provide valuable priors that can guide more effective flight planning.

Given the recent influence of imitation learning in trajectory planning \cite{he_flying_2025}, we employ IL to incorporate the observed trajectory priors into our planning framework.
To promote training efficiency and reduce unnecessary input complexity, we adopt a compact observation space $\boldsymbol{o} := [\boldsymbol{p}_{\text{rel}, xy}, \theta_{\text{inc}}]$, where $\boldsymbol{p}_{\text{rel}, xy}$ encodes the relative position of the quadrotor in the horizontal plane with respect to the inclined surface, and $\theta_{\text{inc}}$ denotes the surface's inclination angle, derived from the surface orientation vector $\boldsymbol{o}_{\text{inc}}$.
The action space comprises local polynomial trajectory parameters near the surface, defined as $\boldsymbol{a} := [\boldsymbol{c}_x, \boldsymbol{c}_y, \boldsymbol{c}_z]$.
This polynomial parameterization enables a compact network architecture to learn trajectory "tendencies" more effectively and stably than approaches based on discrete trajectory points. 

Moreover, for analogical observations, multiple reasonable trajectories may exist. A deterministic network modeling ($\boldsymbol{o} \rightarrow \boldsymbol{a}$) risks learning averaged action patterns that may not represent valid individual trajectories \cite{liu_rdt-1b_2025}. 
Diffusion models can naturally handle this multi-modality by modeling the conditional probability $p(\boldsymbol{a}|\boldsymbol{o})$.
Therefore, we design a Diffusion model to address the multi-modal nature of valid trajectories \cite{peebles_scalable_2023}. The model structure can be seen in \Cref{fig:waypointconstraint_new}.

We collect many high-quality, effective single-task trajectories as training data. For each clean action $\boldsymbol{a}^0$, we perform $K$ Gaussian noise addition steps to obtain noisy action $\boldsymbol{a}^K$:
\begin{equation}
\boldsymbol{a}^k = \sqrt{\bar{\alpha}_k}\boldsymbol{a}^0 + \sqrt{1-\bar{\alpha}_k}\boldsymbol{\epsilon}, \; \boldsymbol{\epsilon} \sim \mathcal{N}(\mathbf{0}, \mathbf{I}), \; k=1,\ldots,K
\end{equation}
where $\bar{\alpha}_k = \prod_{i=1}^k \alpha_i$ and $\alpha_k = 1-\beta_k$ with $\beta_k$ being the noise schedule. The denoising network $f_{\theta}$ is trained to predict the noise:
\begin{equation}
\mathcal{L}(\theta) = \mathbb{E}_{k,\boldsymbol{a}^0,\boldsymbol{\epsilon}} \left[ \| \boldsymbol{\epsilon} - f_{{\theta}}(\boldsymbol{a}^k, k, \boldsymbol{o}) \|^2 \right]
\end{equation}

During inference, we sample from the learned distribution through the reverse process:
\begin{equation}
\boldsymbol{a}^{k-1} = \frac{1}{\sqrt{\alpha_k}} \left( \boldsymbol{a}^k - \frac{\beta_k}{\sqrt{1-\bar{\alpha}_k}}  f_{\theta}(\boldsymbol{a}^k, k, \boldsymbol{o}) \right) + \sigma_k \boldsymbol{z}
\end{equation}
where $\sigma_k = \sqrt{\beta_k}$ and $\boldsymbol{z} \sim \mathcal{N}(\mathbf{0}, \mathbf{I})$. This reverse diffusion process generates diverse trajectory parameters that capture the multi-modal distribution of successful manipulation strategies.

\subsection{IL-guided Optimization}\label{sec:il_guided_optimization}
After learning polynomial trajectories through IL, we leverage the IL prior trajectories as guidance for optimization by incorporating them as quadrotor waypoint constraints. We sample $k$ guide points from the IL prior polynomial trajectory and insert them into \Cref{eq: wpconstraint}:
\begin{equation}
\boldsymbol{w}_{\text{guide}} := \{ \boldsymbol{p}_{\text{guide},1},\ldots,\boldsymbol{p}_{\text{guide},k} \} \subset \mathbf{W}_{\text{cons}}
\end{equation}
where $\boldsymbol{p}_{\text{guide},i}$ represents the $i$-th guide point. 
These sampled guide points ensure that the aerial manipulator follows the desired trajectory direction, thereby improving success rates to some extent.

However, trajectories constrained to fixed guide points may still be suboptimal due to their rigidity.
To address this limitation, we implement a novel two-stage optimization strategy. 
In the first stage, we enforce guide point constraints with a loose gradient tolerance condition $\epsilon_\text{loose}$, which ensures that the optimization is guided toward the desired direction during the initial search phase.
In the second stage, we remove the guide point constraints and apply strict convergence criteria $\epsilon_\text{strict}$, allowing the optimizer to discover locally superior solutions while maintaining the beneficial trajectory structure established in the first stage. 
This two-stage approach combines the global guidance from successful demonstrations with local optimization refinement, significantly improving both success rates and trajectory quality for complex manipulation tasks.
The complete procedure is detailed in \Cref{alg:IL_guided_optimization}.

\section{Experiments}\label{sec: experiments}

This section presents a comprehensive experimental evaluation of the proposed framework across multiple dimensions. We begin by detailing the imitation learning data collection and training procedures for the DiT model. Subsequently, we conduct ablation studies to validate key components: the SFC enlarge strategy and the IL-guided optimization approach. We then benchmark different collision volume approximation methods, comparing static ellipsoid and multi-sphere fitting approaches. Simulation experiments demonstrate the framework's versatility across various manipulator configurations, including telescopic manipulators, 1-DOF, and 2-DOF robotic arms. Finally, real-world experiments validate the system's performance through the execution of 9 fundamental manipulation skills: \textit{strike}, \textit{grasp}, \textit{lift}, \textit{press}, \textit{wind}, \textit{write}, \textit{cross}, \textit{push}, and \textit{pull}, confirming the practical applicability and versatility of our integrated approach.

\subsection{Data Collection and IL Training}\label{sec: training}
For the IL-guided optimization, we focus on grasp and strike skills. Expert trajectories are generated on inclined plane scenarios with angles uniformly sampled from $20^\circ$ to $60^\circ$. Initial robot positions are randomly sampled within annular regions of $1.5$m to $3.0$m radius centered on the inclined plane, with a task-specific waypoint constraint for the end-effector positioned at $0.05$m directly above the plane. Due to low success rates in direct planning, we apply the two-stage optimization algorithm similar to \Cref{sec:il_guided_optimization}. However, the guide points are strategically selected through expert knowledge before and after critical strike and grasp phases. Generated trajectories with total optimization cost exceeding $1 \times 10^5$ are discarded to ensure solution quality.

To obtain stable trajectory tendencies within local regions, we extract trajectory segments of $\pm 1$s duration around the grasping and striking phases. From each segment, we sample 15 discrete points that are subsequently fitted using a three-dimensional fifth-order polynomial $[\boldsymbol{c}_x, \boldsymbol{c}_y, \boldsymbol{c}_z]$. The resulting 15 polynomial coefficients constitute the trajectory tendency representation that serves as training targets for the diffusion model. Through this process, we collected approximately 3,000 high-quality trajectory samples for each skill type (grasping and striking) along with their corresponding polynomial representations.
The data collection process requires an average generation time of about 33s per sample, which allows for practical data collection at scale.

The imitation learning component employs a Diffusion Transformer (DiT) architecture specifically designed for trajectory tendency learning. The model utilizes a 128-dimensional embedding space with 8 attention heads across 6 transformer layers, incorporating adaptive layer normalization and cross-attention mechanisms for effective condition-context fusion. With approximately 3M parameters, the network maps 3-dimensional input conditions (relative position $\boldsymbol{p}_{\text{rel},xy}$ and inclination angle $\theta_\text{inc}$) to 15-dimensional polynomial coefficients representing trajectory tendencies. Training is conducted on an NVIDIA RTX 4090 with 24GB memory for approximately one hour using mixed precision and a learning rate of $3 \times 10^{-4}$ with cosine annealing.

% Examples of the model-generated polynomial fits are shown in \Cref{fig:training}.

% \begin{figure}[t]
%     \centering
%     \includegraphics[width=1\linewidth]{assets/training.png}
%     \caption{Model-generated polynomial fits for striking (top three rows) and grasping (bottom three rows) trajectories. Each skill type shows three sampled outputs from the trained diffusion model. Star symbols indicate starting positions, green rectangles represent ramp locations with tilt angles labeled above, and colored curves show the fitted polynomial trajectories.}
%     \label{fig:training}
% \end{figure}

\begin{figure}[t]
    \centering
    \includegraphics[width=1\linewidth]{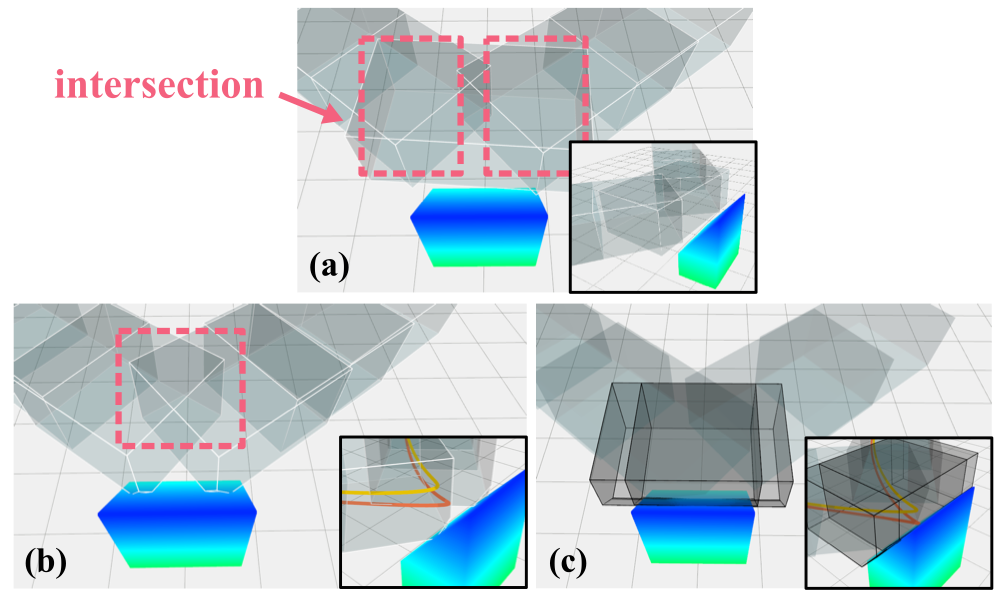}
    \caption{Comparison of different polyhedron generation methods for inclined surface grasping: (a) original method from \cite{lee_planning_2018} without modifications, (b) ablated method with manually constrained intersection above the inclined surface, and (c) our proposed Active Polyhedron Generation method. The pink boxes indicate the locations of adjacent polyhedron intersections.}
    \label{fig:sfc_ablation}
\end{figure}

\begin{table}[t]
\centering
\caption{Quantitative comparison of Active Polyhedron Generation methods}
\label{tab:sfcablation}
\renewcommand{\arraystretch}{1.2}
\begin{tabular*}{\linewidth}{@{\extracolsep{\fill}}lccc@{}}
\toprule
\textbf{Method} & \textbf{Utilization Rate (\%)} & \textbf{Final Cost} & \textbf{Distance (m)} \\
\midrule
\textbf{Ablated} & 27.1 & 9.51 $\times$ 10$^5$ & 0.267 \\
\textbf{Proposed} & 100.0 & 3.54  $\times$ 10$^3$ & 0.027 \\
\bottomrule
\end{tabular*}
\end{table}

\subsection{Ablation Study}

We designed two ablation experiments to validate the effectiveness of our Active Polyhedron Generation and demonstrate the improvement provided by IL-guided Optimization.

\textbf{1) Active Polyhedron Generation}: To verify that our Active Polyhedron Generation not only maintains compatibility with our waypoint constraint framework but also expands the solution space, we designed an inclined surface grasping scenario with three different configurations: (1) the original method from \cite{lee_planning_2018} without any modifications; (2) a modified version that ensures the intersection of two polyhedrons exists directly above the inclined surface to enable waypoint constraint application; and (3) our proposed Active Polyhedron Generation method.

The results are shown in \Cref{fig:sfc_ablation}. In \Cref{fig:sfc_ablation} (a), without predetermined intervention points in SFC generation, the resulting SFC exhibits significant randomness. The polyhedron intersections are located on both sides of the inclined surface, making it difficult to effectively apply waypoint constraints. In \Cref{fig:sfc_ablation} (b), although we constrained the intersection to exist directly above the inclined surface, substantial unused space remains, particularly near the center of the inclined surface, resulting in the "hollow" region we mentioned. In contrast, \Cref{fig:sfc_ablation} (c) demonstrates our proposed Active Polyhedron Generation, which actively generates polyhedrons that nearly achieve full utilization of the feasible space above the inclined surface.

Since configuration (1) cannot accommodate waypoint constraints, we conducted quantitative analysis only on configurations (2) and (3). We applied our optimization framework to optimize the grasp skill trajectory, with results shown in the lower panels of \Cref{fig:sfc_ablation} (a) and (c), where orange trajectories represent the end-effector path and yellow trajectories represent the quadrotor path. Our proposed method successfully reaches the target, while the ablated method suffers from conflicts between end-effector waypoint constraints and safety constraints due to limited solution space. This leads to failure in reaching the target point, and excessive safety penalties in the vicinity result in significantly degraded trajectory quality.

\Cref{tab:sfcablation} presents a quantitative comparison of three key metrics. The Utilization Rate  measures the percentage of free space coverage within a 0.5m radius sphere around the target point. The Final Cost represents the objective function value after convergence. The Distance quantifies the end-effector positioning error at the waypoint constrained location.

\begin{figure}
    \centering
    \includegraphics[width=1\linewidth]{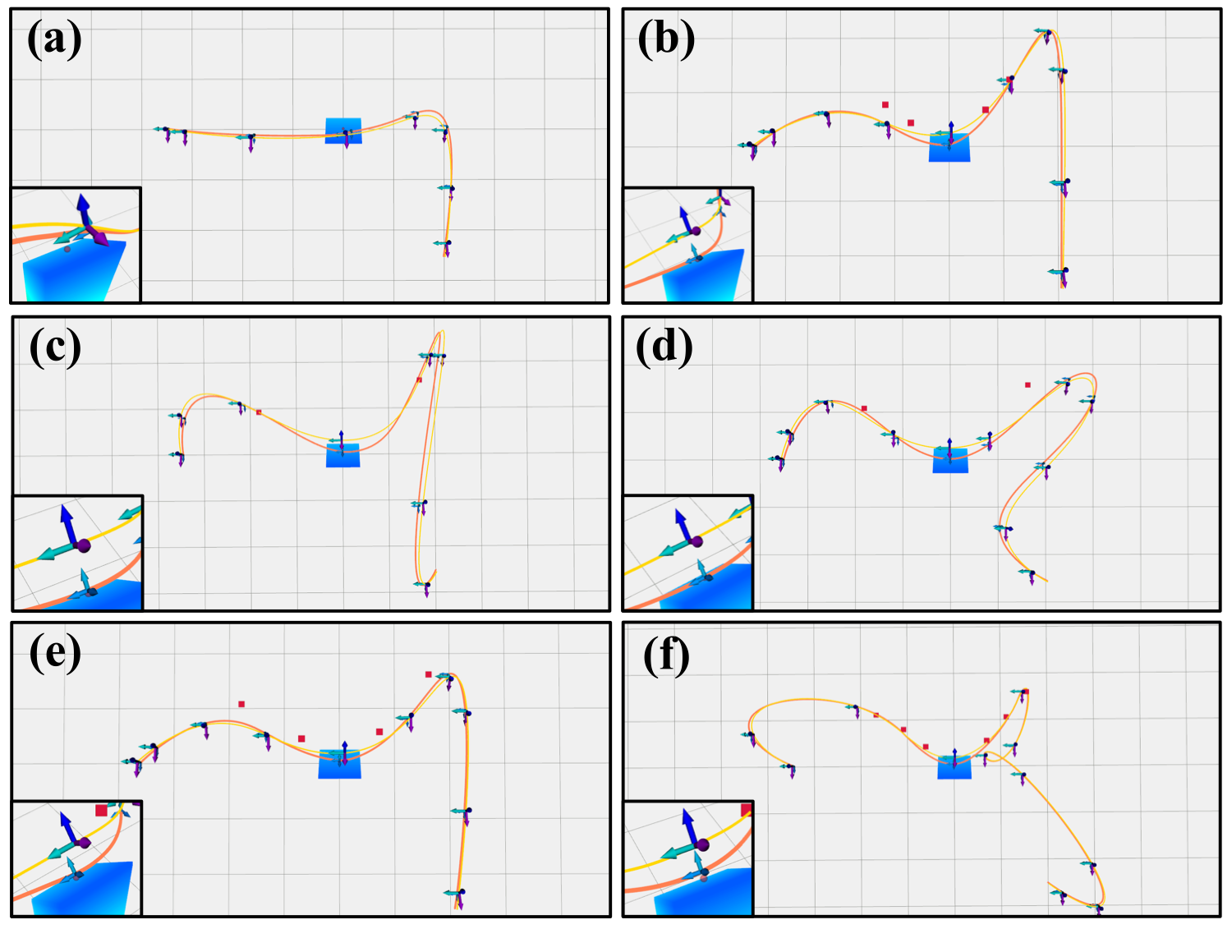}
    \caption{Convergence results for IL-guided optimization variants. Yellow trajectories represent the motion of the quadrotor, while orange trajectories show the end-effector’s movement in $\mathcal{F}_\mathcal{W}$. Red points indicate IL-derived guide points, and aerial manipulator orientations at strike instant are shown in the lower-left corners.}
    \label{fig:IL_guided}
\end{figure}

\begin{figure}
    \centering
    \includegraphics[width=1\linewidth]{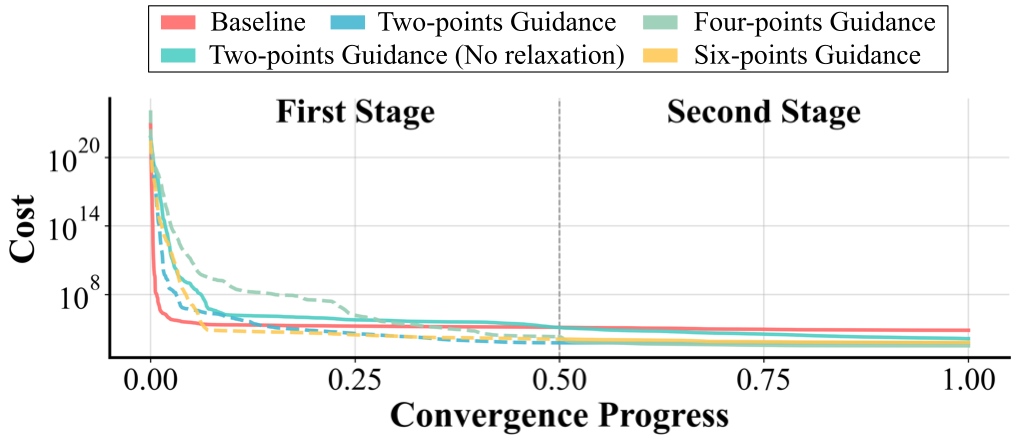}
    \caption{Convergence curves comparison across optimization methods. Normalized iteration lengths account for varying convergence requirements. Dashed lines: first-stage optimization; solid lines: second-stage optimization.}
    \label{fig:convergence}
\end{figure}

% \begin{figure}
%     \centering
%     \includegraphics[width=0.5\linewidth]{assets/converage.png}
%     \caption{Enter Caption}
%     \label{fig:placeholder}
% \end{figure}

% \begin{figure}
%     \centering
%     \includegraphics[width=0.5\linewidth]{assets/convergence_curves.png}
%     \caption{Enter Caption}
%     \label{fig:placeholder}
% \end{figure}

\begin{table*}[ht]
\centering
\caption{Comparison of Robotic Manipulation Optimization Methods Across Different Roll Angles}
\label{tab:method_comparison}
\begin{tabular*}{\textwidth}{@{\extracolsep{\fill}}llccccccc@{}}
\toprule
\multirow{2}{*}{\textbf{Method}} & \multirow{2}{*}{\textbf{Roll (${}^\circ$)}} & \multicolumn{2}{c}{\textbf{Performance}} & \multicolumn{3}{c}{\textbf{Trajectory Quality}} & \multicolumn{2}{c}{\textbf{Efficiency}} \\
\cmidrule(lr){3-4} \cmidrule(lr){5-7} \cmidrule(lr){8-9}
& & \textbf{Success Rate (\%)} & \textbf{Cost} & \textbf{Time (s)} & \textbf{Length} & \textbf{Jerk} & \textbf{Iteration} & \textbf{Runtime (s)} \\
\midrule
\multirow{4}{*}{\shortstack[l]{Baseline \\ (Complete \\ Ablation)}} 
& 25 & \textbf{100} & \textbf{2137.4} & \textbf{3.7} & \textbf{6.2} & \textbf{277.7} & \textbf{1536} & \textbf{5.1} \\
& 40 & 50$^{\dagger}$ & 4541.7 & 6.8 & 9.6 & \textbf{281.1} & \textbf{1726} & \textbf{6.1} \\
& 55 & 0$^{\dagger}$ & --- & --- & --- & --- & --- & --- \\
& 40(OOD) & 70 & 3703.1 & 4.6 & \textbf{8.5} & 488.2 & \textbf{2619} & \textbf{7.9} \\
\midrule
\multirow{4}{*}{\shortstack[l]{Learned Points \\ Guidance + \\ Relaxation}} 
& 25 & 90 & 3903.5 & 6.6 & 8.1 & 338.5 & 34306$^{\dagger}$ & 132.4 \\
& 40 & 70 & 5527.5$^{\dagger}$ & 10.1$^{\dagger}$ & 9.7$^{\dagger}$ & 400.8 & 37209 & 139.4$^{\dagger}$ \\
& 55 & 80 & 5127.2 & 8.2 & \textbf{11.1} & 886.0 & 19653 & 86.5 \\
& 40(OOD) & 60 & 4081.4& 6.2 & 8.8 & 440.6 & 47118$^{\dagger}$ & 151.2$^{\dagger}$ \\
\midrule
\multirow{4}{*}{\shortstack[l]{{Two-Point} \\ {Guidance} \\ {(No Relaxation)}}} 
& 25 & \textbf{100} & 3994.9$^{\dagger}$ & 5.8 & 8.0 & 401.5$^{\dagger}$ & 8272 & 31.7 \\
& 40 & 70 & 3243.0 & 4.7 & \textbf{8.0} & 579.1$^{\dagger}$ & 10653 & 35.4 \\
& 55 & \textbf{100} & \textbf{4137.0} & \textbf{6.1} & 12.6 & 974.4$^{\dagger}$ & \textbf{5590} & \textbf{21.9} \\
& 40(OOD) & 90 & 3035.4 & 4.9 & 9.7 & 580.1 & 6904 & 27.2 \\
\midrule
\multirow{4}{*}{\shortstack[l]{\textbf{Two-Point} \\ \textbf{Guidance +} \\ \textbf{Relaxation} \\ \textbf{(Ours)}}} 
& 25 & \textbf{100} & 2568.9 & 4.4 & 6.8 & 317.2 & 9364 & 34.6 \\
& 40 & \textbf{100} & 3584.1 & 5.4 & 9.0 & 563.1 & 16357 & 50.9 \\
& 55 & \textbf{100} & 4612.0 & 7.3 & 13.2$^{\dagger}$ & \textbf{857.1} & 6741 & 27.4 \\
& 40(OOD) & \textbf{100} & \textbf{2719.2} & \textbf{4.2} & 8.8 & 584.9$^{\dagger}$ & 9566 & 35.6 \\
\midrule
\multirow{4}{*}{\shortstack[l]{\textbf{Four-Point} \\ \textbf{Guidance +} \\ \textbf{Relaxation} \\ \textbf{(Ours)}}} 
& 25 & 80 & 2842.3 & 4.4 & 7.3 & 377.8 & 22529 & 88.2 \\
& 40 & 70 & \textbf{3112.3} & \textbf{5.0} & 8.5 & 575.8 & 37550 & 126.2 \\
& 55 & 70 & 4376.9 & 6.9 & 11.9 & 852.5 & 17751 & 88.4 \\
& 40(OOD) & 90 & 4053.7$^{\dagger}$ & 7.1$^{\dagger}$ & 11.4$^{\dagger}$ & 446.8 & 7460 & 34.0 \\
\midrule
\multirow{4}{*}{\shortstack[l]{\textbf{Six-Point} \\ \textbf{Guidance +} \\ \textbf{Relaxation} \\ \textbf{(Ours)}}} 
& 25 & 70$^{\dagger}$ & 3838.8 & 6.9$^{\dagger}$ & 8.6$^{\dagger}$ & 325.7 & 33666 & 172.9$^{\dagger}$ \\
& 40 & 50 & 4250.7 & 6.8 & 9.1 & 519.4 & 56956$^{\dagger}$ & 239.6 \\
& 55 & 60 & 6097.0$^{\dagger}$ & 10.5$^{\dagger}$ & 12.6 & 737.4 & 23621$^{\dagger}$ & 146.0$^{\dagger}$ \\
& 40(OOD) & 30$^{\dagger}$ & 3389.2 & 5.9 & 9.3 &\textbf{406.6} & 37985 & 149.3 \\
\bottomrule
\end{tabular*}%
\vspace{0.1cm}
\begin{tablenotes}
\footnotesize
\item \textbf{Bold}: Best performance within each roll \quad $^{\dagger}$ Worst performance within each roll \quad OOD: Given out-of-distribution initial states
\end{tablenotes}
\end{table*}

\textbf{2) IL-guided Optimization}: To validate the effectiveness of our IL-guided optimization framework, we conduct a comprehensive ablation study examining strike success rates on inclined surfaces. This study addresses three key questions: whether our IL-guided optimization can improve strike success rates, the rationale for sampling points from polynomial trajectories rather than learning discrete points directly, and the optimal number of sampling points for trajectory guidance.

We design the experimental scenario with varying inclined surfaces and designated target points for strike operations. A successful strike requires two simultaneous conditions at contact: the aerial manipulator maintains parallel alignment with the inclined surface, and the end-effector remains within $0.03$m of the target center. Initial positions are randomly sampled within a $1.5-3$m annular region around the surface for training, while out-of-distribution (OOD) testing uses a $3-4$m annular region to evaluate generalization performance. We assume prior knowledge of target location and surface inclination angle, establishing fixed end-effector waypoint constraints and corresponding orientations. Our experimental design comprises six configurations tested across four scenarios: 25$^{\circ}$, 40$^{\circ}$, and 55$^{\circ}$ surface angles with in-distribution initial positions, 40° with OOD initial positions. 

The \textbf{baseline} completely removes IL-guided optimization, relying solely on direct trajectory optimization. The second configuration (\textbf{Learned Points Guidance + Relaxation}) trains the IL model to predict four discrete points, maintaining consistency with our training methodology but reducing the projection dimension from $15$ to $12$. The third approach (\textbf{Two-Point Guidance, No Relaxation}) implements our polynomial trajectory imitation learning, sampling one point before and after strike as guide points with strict convergence criteria. 
Our primary method (\textbf{Two-Point Guidance + Relaxation}) extends the third configuration by incorporating relaxed convergence criteria and two-stage optimization, removing guide points in the second stage for stricter optimization. 
Finally, we evaluate \textbf{Four-Point and Six-Point Guidance + Relaxation} variants, differing only in the number of polynomial trajectory sampling points.

\Cref{fig:IL_guided} illustrates convergence results for challenging cases where initial positions lie behind the inclined surface. The baseline approach in \Cref{fig:IL_guided} (a) demonstrates complete failure, with the end-effector trajectory (orange) missing the target entirely due to convergence to poor local optima. In contrast, learned point guidance in \Cref{fig:IL_guided} (b) successfully captures required trajectory tendencies, achieving target strikes. Our two-point sampling approach \Cref{fig:IL_guided} (c) produces quadrotor trajectories (yellow) that accurately traverse the red guide points, resulting in desired trajectory generation. Both two-point and four-point guidance \Cref{fig:IL_guided} (d) and (e) consistently accomplish the task, while six-point guidance \Cref{fig:IL_guided} (f) suffers from over-constraint, leading to suboptimal solutions. \Cref{fig:convergence} quantitatively demonstrates that all proposed IL-guided methods achieve superior local optima compared to baseline optimization, as evidenced by their lower convergence costs. Single-stage optimization outperforms baseline but remains inferior to two-stage approaches, confirming that the relaxation strategy further enhances local optimum extraction by allowing more effective exploration in the first stage followed by refined convergence in the second stage.

\Cref{tab:method_comparison} presents comprehensive quantitative analysis across performance metrics, trajectory quality, and computational efficiency. Our Two-Point Guidance + Relaxation achieves \textbf{100\%} success rates across all scenarios, significantly outperforming alternatives, particularly in the challenging $55^{\circ}$ case where baseline success rate drops to zero. This demonstrates the critical importance of IL guidance for steep surface manipulation. Two-Point Guidance without relaxation ranks second, while Four-Point and Six-Point Guidance exhibit reduced success rates due to over-constraint in the first optimization stage, where two points already provide sufficient information for trajectory guidance, and additional points excessively constrain the trajectory during the first stage, thereby compromising the second stage's search capability.

Regarding generalization performance, while Learned Points Guidance + Relaxation achieves comparable success rates in standard scenarios, it significantly underperforms in OOD conditions. This performance gap stems from polynomial fitting's superior ability to integrate multi-point information, producing more stable results compared to discrete point prediction, which suffers from data sparsity as identical trajectories generate numerous possible point combinations. Additionally, Learned Points Guidance + Relaxation exhibits significantly longer convergence times and more iteration times compared to our proposed methods, indicating that the learned points deviate substantially from the desired optimal trajectory, further highlighting the deficiencies of this approach.

Computational analysis reveals that increasing guide points extends convergence time compared to baseline performance, as guide points direct optimization toward potentially distant desired directions, which increases the iteration requirements. This indicates that setting more guide points leads to increased solution time and complexity for the optimization problem.

Interestingly, Two-Point Guidance + Relaxation occasionally produces higher final costs than the no-relaxation variant, reflecting optimization randomness where relaxed first-stage criteria may lead to different local optima. However, cost differences remain minimal, which is still acceptable. 
In terms of trajectory quality, we observe that IL successfully learns high-quality trajectory tendencies from demonstration data. Furthermore, increasing the number of guide points yields lower final jerk values, indicating that more guide points drive the optimization closer to these learned high-quality trajectories.

These findings collectively demonstrate that our IL-guided optimization framework successfully enhances performance, with our polynomial approach exhibiting superior stability and generalization compared to learned point methods. Moreover, selecting merely two points can provide sufficient trajectory tendency information while maintaining computational efficiency.

% \begin{figure*}[ht]
%     \centering
%     \includegraphics[width=1\linewidth]{assets/Ablations.png}
%     \caption{Experimental validation of our trajectory planning framework with dynamic approximation. (a) shows the swept volume under dynamic approximation in the Narrow Gate scenario. (b) illustrates the ablation study, clearly showing that without our dynamic approximation module, the delta arm fails to retract and results in collision. (c) and (d) demonstrate the effectiveness of our method even under large attitude variations. (e) and (f) showcase that only our proposed method successfully navigates through narrow gaps in random cube environments without collisions.}
%     \label{fig:ablation}
% \end{figure*}

\begin{figure}
    \centering
    \includegraphics[width=1\linewidth]{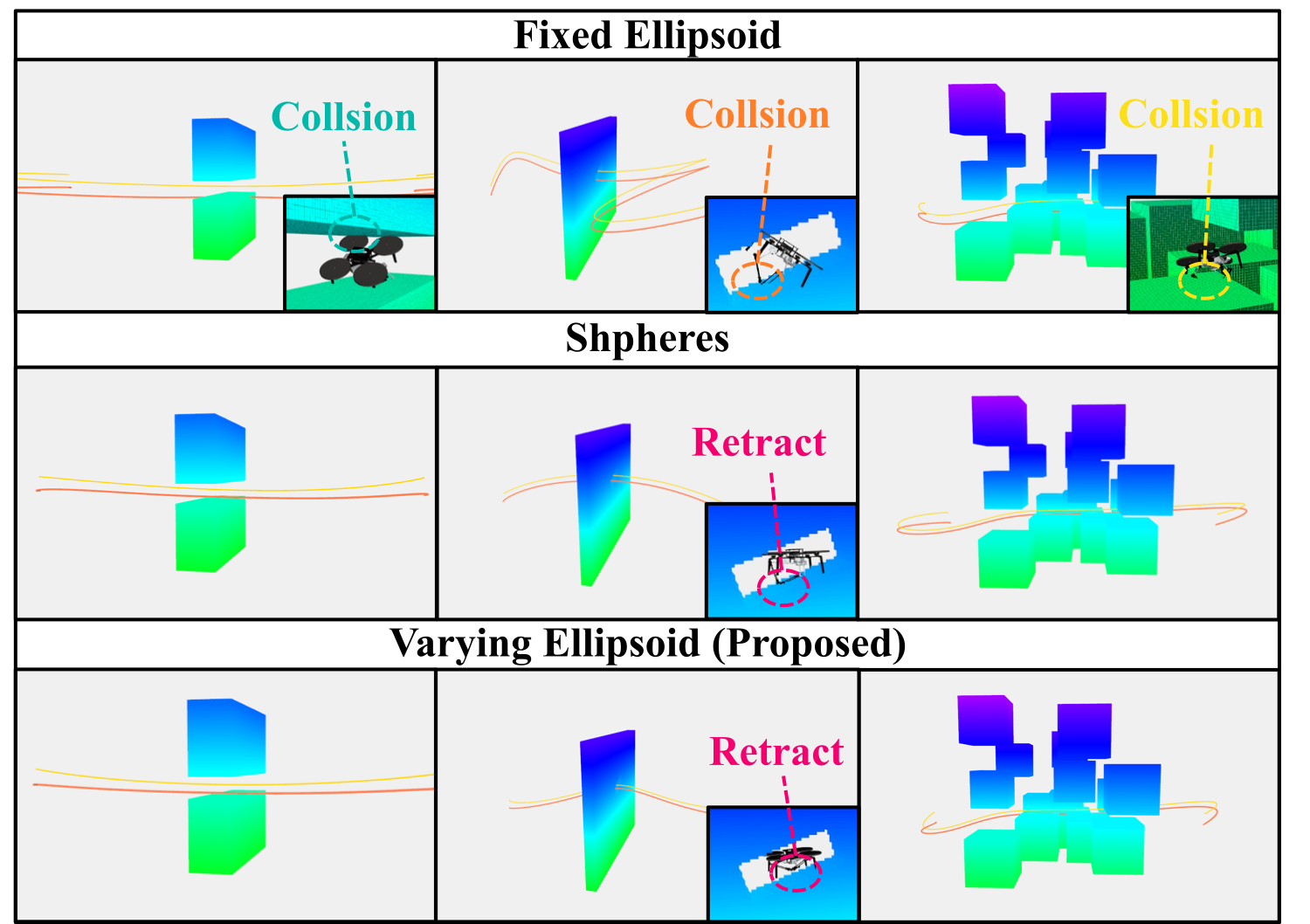}
    \caption{Collision avoidance performance comparison across three benchmark scenarios: narrow gate (left), tilted hole (middle), and random cubes (right). The fixed ellipsoid method (top row) results in collisions due to rigid volume representation. The spheres method (middle row) achieves collision-free navigation. Our varying ellipsoid method (bottom row) demonstrates superior performance through adaptive manipulator retraction.}
    \label{fig:collision_benchmark}
\end{figure}

\subsection{Benchmark}
To validate the superiority of our varying ellipsoid method for collision volume approximation, we designed three obstacle-rich scenarios and evaluated the aerial manipulator's collision avoidance performance through each environment. 

The first scenario, termed "\textbf{Narrow Gate}", consists of two rectangular obstacles positioned to create a narrow passage between them, as illustrated in the first column of \Cref{fig:collision_benchmark}. We systematically varied the gap width from $0.60$m to $0.25$m in $0.05$m decrements, resulting in eight experimental trials. Throughout these experiments, the aerial manipulator was manually configured to maintain an extended state at both the initial and final positions, with the end-effector position set to $^{\mathcal{D}}p_{e,z} = -0.2$m to ensure consistent testing conditions.
The second scenario, "\textbf{Tilted Hole}," features a wall with an inclined hole, as shown in the second column of \Cref{fig:collision_benchmark}. We examined three different inclination angles: $20^{\circ}$, $40^{\circ}$, and $60^{\circ}$. For each angle, the manipulator's initial end-effector positions were set to $^{\mathcal{D}}p_{e,z} = -0.07$m, $-0.13$m, and $-0.2$m respectively, yielding a total of nine experimental configurations.
The third scenario, "\textbf{Random Cubes}", involves randomly distributing $12$ cubic obstacles within a three-dimensional workspace. Ten different random configurations were generated to provide diverse navigation challenges for the aerial manipulator. For all experiments, a trial was considered successful if the aerial manipulator reached the target position without any collision occurrence during the entire trajectory.

We compared our approach against two established collision volume approximation methods: Fast-Racing~\cite{han_fast-racing_2021} and REMANI~\cite{wu_real-time_2024}. The Fast-Racing method employs a single fixed ellipsoid to approximate the entire aerial manipulator system. To ensure collision safety, this fixed ellipsoid must be sized to accommodate the manipulator in its fully extended configuration. The REMANI method utilizes multiple spheres to represent the aerial manipulator's geometry: four spheres with 0.08m radius for the quadrotor body, one sphere with 0.025m radius for the end-effector, and fifteen spheres with 0.025m radius distributed along the upper and lower arm segments. All experimental comparisons maintained identical constraint weights and hyperparameters across the 27 total trials.

The results presented in \Cref{fig:collision_benchmark} demonstrate clear performance distinctions between methods. The fixed ellipsoid approach failed to navigate collision-free through gaps smaller than or equal to $0.45$m, as the rigid geometric representation prevented adaptive arm retraction during narrow passage traversal. Similarly, this method failed in the tilted hole scenario when the initial end-effector positions were set to $-0.13$m and $-0.2$m. The random cubes scenario also resulted in collisions when narrow passages were encountered.

As summarized in \Cref{tab:obstacle_avoidance_comparison}, the fixed ellipsoid method achieved only $37.5\%$ and $33.3\%$ success rates in the first two scenarios, respectively, while maintaining $70\%$ success in the random cubes scenario due to the relatively sparse obstacle distribution.

In contrast, both the sphere-based method and our proposed varying ellipsoid approach achieved $100\%$ success rates in the narrow gate and tilted hole scenarios. However, their collision avoidance strategies differed significantly, as observable in the bottom-right panel of the second column in \Cref{fig:collision_benchmark}. The sphere-based method tends to induce lateral ($xy$-axis) manipulator movements, while our approach primarily achieves collision avoidance through strategic end-effector retraction along the $z$-axis.

For the random cubes scenario, our proposed method achieved an $80\%$ success rate compared to $70\%$ for the sphere-based approach. The failures in both methods were primarily attributed to path planning limitations that forced the aerial manipulator into excessively constrained regions where sufficient SFC volume could not be established. Additionally, the sphere-based method's failures resulted from the cumulative high costs generated by multiple spheres in proximity to obstacles, preventing the optimizer from adequately balancing other essential constraints and leading to localized collisions.

The computational efficiency comparison in \Cref{tab:obstacle_avoidance_comparison} reveals that our method's average iteration time closely matches that of the fixed ellipsoid approach while significantly outperforming the sphere-based method. This efficiency stems from our approach's $O(n)$ computational complexity, as only a single ellipsoid-to-polyhedron cost evaluation is required per iteration, compared to the $O(mn)$ complexity of the sphere-based method that necessitates individual cost calculations for each sphere.

In summary, our varying ellipsoid fitting method successfully combines the computational efficiency of ellipsoidal representations with comprehensive whole-body collision avoidance capabilities for aerial manipulators.

\begin{table}[tbp]
\centering
\caption{Performance Comparison of Obstacle Avoidance Methods}
\label{tab:obstacle_avoidance_comparison}
\resizebox{\columnwidth}{!}{%
\begin{tabular}{lcccc}
\toprule
\textbf{Method} & \textbf{Narrow Gate} & \textbf{Tilted Hole} & \textbf{Random Cubes} & \textbf{Avg. Iteration Time} \\
\midrule
Fast-Racing~\cite{han_fast-racing_2021} & 37.5\% & 33.3\% & 70.0\% & 33.85\,ms \\
\addlinespace[0.3em]
REMANI~\cite{wu_real-time_2024} & \textbf{100.0\%} & \textbf{100.0\%} & 70.0\% & 123.73\,ms \\
\addlinespace[0.3em]
Proposed & \textbf{100.0\%} & \textbf{100.0\%} & \textbf{80.0\%} & \textbf{45.85\,ms} \\
\bottomrule
\end{tabular}%
}
\end{table}

\begin{figure}[t]
    \centering
    \includegraphics[width=1\linewidth]{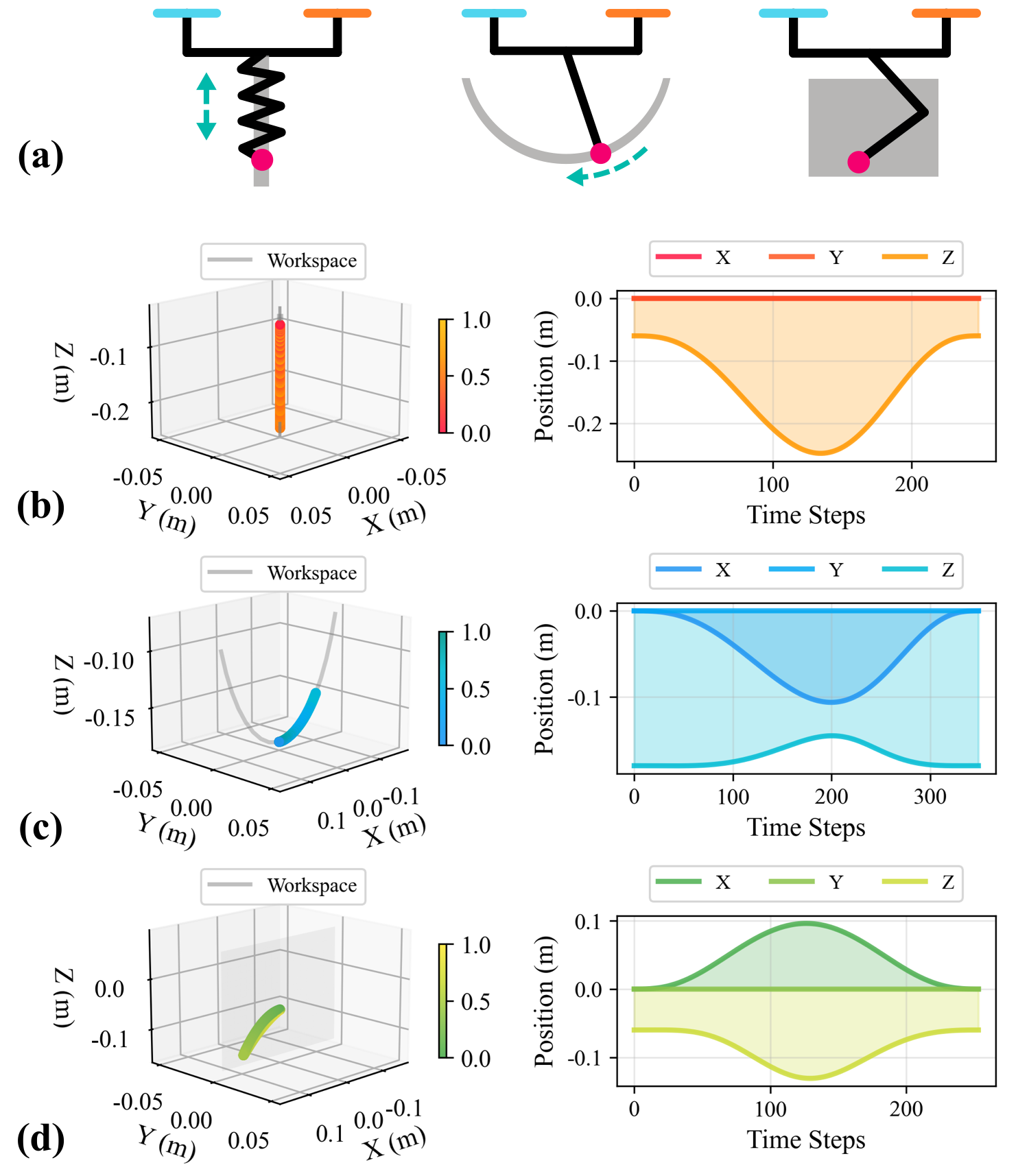}
    \caption{Validation of the planning framework across different manipulator configurations. (a) Three manipulator types: vertical motion (top), 1-DOF circular arc motion (middle), and 2-DOF planar motion (bottom). Left column shows 3D end-effector trajectories within their respective workspaces (gray regions). Right column displays the corresponding position profiles for each axis over time. All trajectories remain strictly within workspace boundaries, demonstrating the framework's adaptability to various kinematic constraints.}
    \label{fig:various_workspace}
\end{figure}

\subsection{Simulation}

We formulate the end-effector planning in Cartesian space specifically to accommodate different aerial manipulator configurations. This section validates that our approach can work with various manipulator types through comprehensive simulation studies.

Due to hardware limitations, we evaluate the feasibility of our approach through simulation analysis. As illustrated in \Cref{fig:various_workspace} (a), we consider three distinct manipulator configurations with different kinematic constraints. The first type is a telescopic manipulator that permits only vertical ($z$-axis) motion of the end-effector. The second type is a 1-DOF manipulator that allows the end-effector to move along a circular arc in the $xz$-plane. The third type is a 2-DOF manipulator that enables planar motion within the entire $xz$-plane. 
The respective workspaces for each configuration are visualized as gray regions in the left column of \Cref{fig:various_workspace}.

We designed a planar grasping scenario with a waypoint constraint for the end-effector in the world frame $\mathcal{F}_\mathcal{W}$, setting the terminal velocity to zero at the target position. For the telescopic manipulator, the workspace bounds were set to $^{\mathcal{D}}p_{e,z} \in [-0.25, -0.06]$m. For the 1-DOF manipulator, the workspace is constrained to a circular arc with radius 0.18m in the $xz$-plane. For the 2-DOF planar manipulator, the workspace was defined as $^{\mathcal{D}}p_{e,x} \in [-0.2, 0.2]$m and $^{\mathcal{D}}p_{e,z} \in [-0.25, -0.06]$m.

The trajectory planning results demonstrate high accuracy, with minimum distances to the target point of $6.851 \times 10^{-3}$m, $4.857 \times 10^{-3}$m, and $2.055 \times 10^{-3}$m for the three manipulator types, respectively. These errors are sufficiently small to confirm successful target reaching. The 3D end-effector trajectories in the frame $\mathcal{F}_\mathcal{D}$, shown in the left column of \Cref{fig:various_workspace}, strictly adhere to their respective workspace constraints. The right column presents the individual axis variations over time, further confirming that all planned trajectories remain within the prescribed workspace boundaries.

These results demonstrate that our whole-body planning framework successfully adapts to different manipulator types while maintaining workspace compliance and trajectory accuracy.

\begin{figure}[t]
   \centering
   \includegraphics[width=1\linewidth]{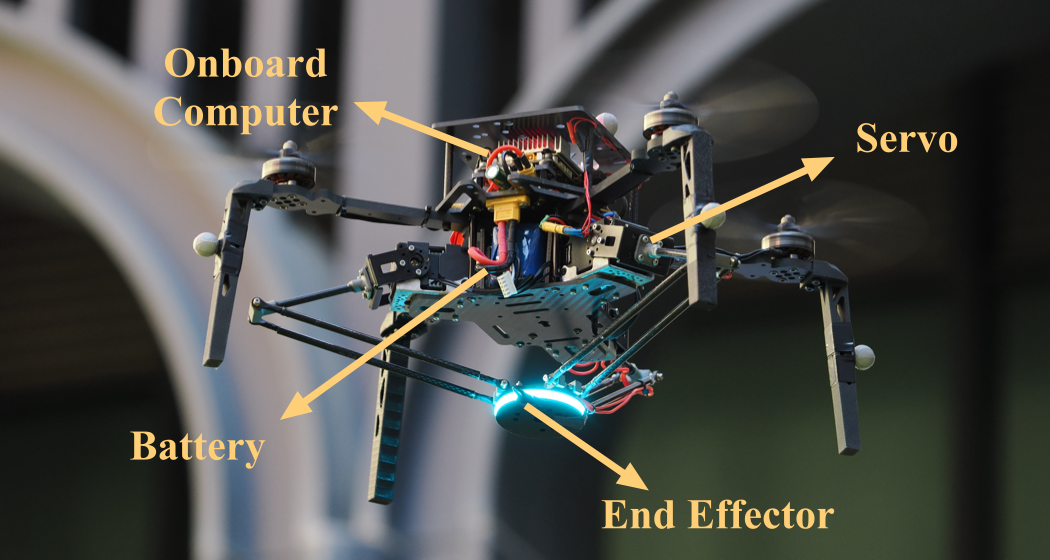}
   \caption{Real-world aerial manipulator platform used for validation.}
   \label{fig:hardware}
\end{figure}

\begin{figure*}[t]
    \centering
    \includegraphics[width=1\linewidth]{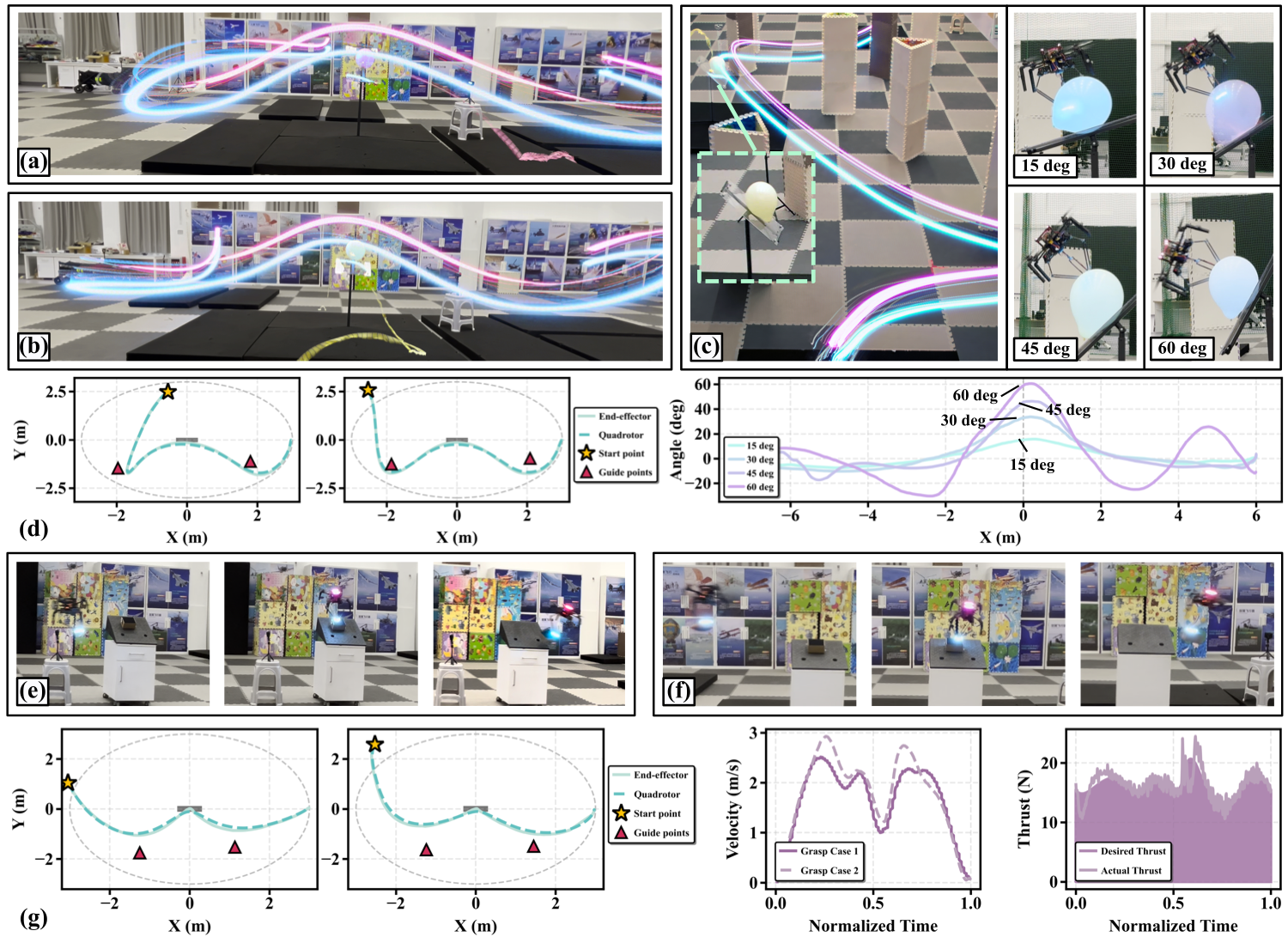}
    \caption{Experimental results for strike and grasp tasks. (a)-(c) show aerial striking experiments while (e) and (f) present aerial grasping results. (d) The first two plots demonstrate IL-generated guide points and resulting quadrotor and end-effector motion in the xy-plane. The third plot shows roll angle variations across four strike maneuvers. (g) The first two plots show IL guide point generation for grasp skills. The third and fourth plots present quadrotor velocity and comparison between planned and actual thrust curves for the aerial manipulator.}
    \label{fig:strikeandstrike}
\end{figure*}

\begin{figure*}
    \centering
    \includegraphics[width=0.95\linewidth]{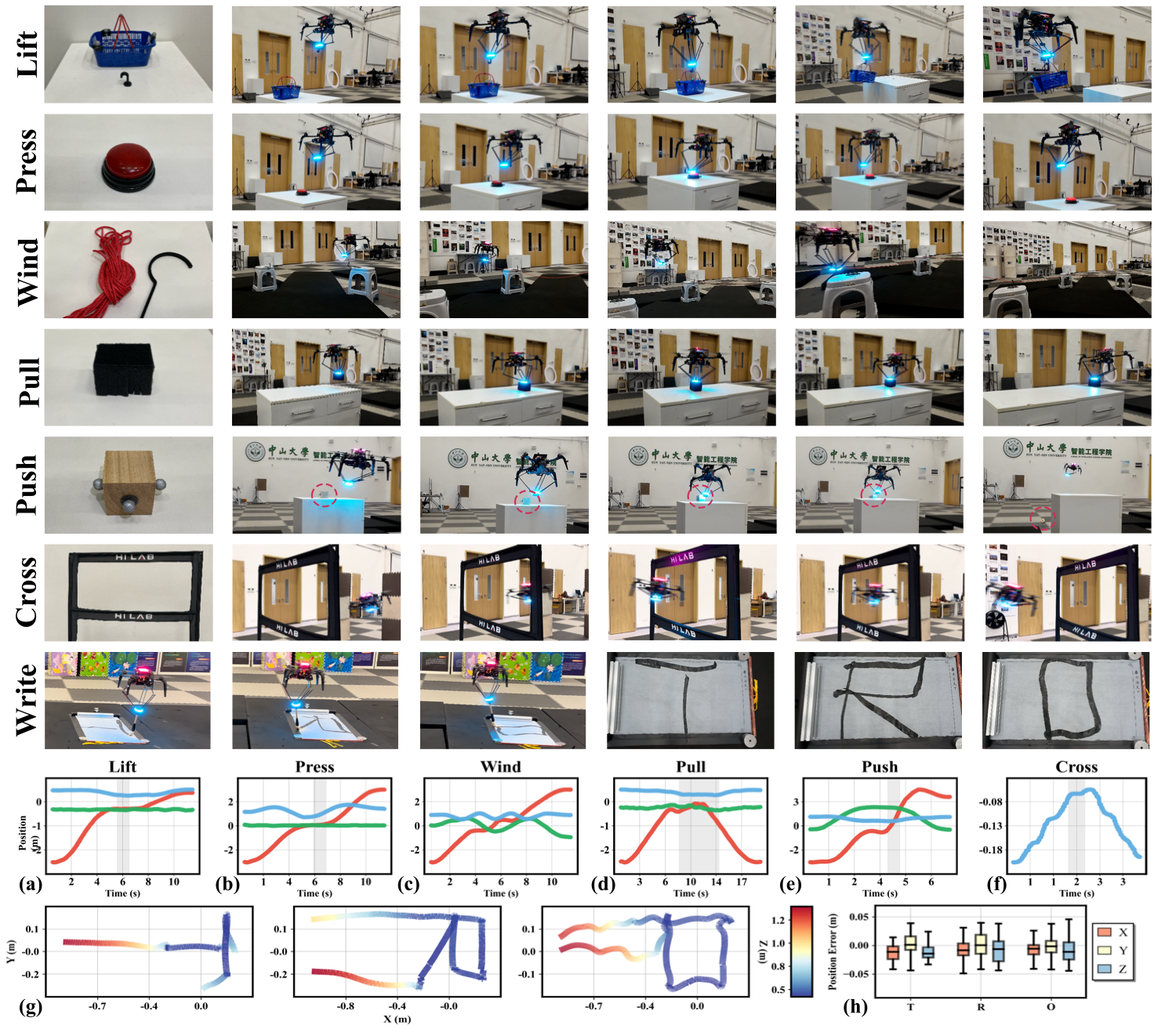}
    \caption{Real flight results for seven tasks: lift, press, wind, pull, push, cross, and write. (a)-(e) show xyz trajectory variations in the world frame for the first five tasks. (f) presents end-effector z-axis motion trajectory in $\mathcal{F}_\mathcal{D}$ during the cross task. (g) and (h) show actual end-effector movement during the write task and corresponding error distributions, respectively.}
    \label{fig:multiple_tasks}
\end{figure*}

\subsection{Real-world Implementation}

Our system is implemented in C++11 on Ubuntu 20.04 with ROS Noetic. As illustrated in \Cref{fig:hardware}, the experimental platform integrates an NVIDIA Jetson Orin NX 16GB as the onboard computer and an NxtPX4v2 flight controller on a quadrotor equipped with a delta arm. The delta arm utilizes three DYNAMIXEL XL430-W250-T servo motors and features a needle end-effector for precise manipulation tasks. For state estimation, the system employs an Extended Kalman Filter (EKF) that fuses data from the NOKOV Motion Capture System and an onboard IMU. 
We implement a separated control framework inspired by the composite control scheme in \cite{chen_ndob-based_nodate}, which employs an NDOB to compensate for internal coupling and external disturbances, while using a high-pass filter to enhance manipulator agility.
Throughout the experimental validation, we assume that interaction forces including gravity and friction are constant and known during the planning phase.

To validate our approach in real-world scenarios, we focus on a fundamental question: whether our flexible waypoint constraints can effectively operate in practical environments and successfully execute diverse manipulation skills. To address this question systematically, we design 9 fundamental manipulation skills: striking, grasping, lifting, pressing, winding, pulling, pushing, crossing, and writing. We conduct a total of 17 experiments across these skills and analyze their performance comprehensively.

\textbf{1) Strike:} We design a striking scenario with an inclined surface positioned at approximately 37.6 degrees at the center of the test area. To thoroughly validate our IL-guided optimization framework, we position the aerial manipulator behind the inclined surface and conduct experiments from two distinct initial positions: one within the training data distribution (1.5-3m radius) and another outside this distribution (beyond 3m radius). The strike skill requires setting a single end-effector waypoint constraint in $\mathcal{F}_\mathcal{W}$ along with its corresponding orientation.

As shown in \Cref{fig:strikeandstrike} (a) and (b), given the initial relative position and the surface inclination angle, our IL-guided framework generates guide points as demonstrated in \Cref{fig:strikeandstrike} (d) (first two plots). To further validate the orientation capabilities of our waypoint constraints, we test the aerial manipulator's striking performance on four different inclined surfaces with angles of 15$^{\circ}$, 30$^{\circ}$, 45$^{\circ}$, and 60$^{\circ}$. To eliminate the influence of other constraints and focus solely on orientation variations, we specify that the delta arm should extend to $^{\mathcal{D}}\boldsymbol{p}_e = [0.0, 0.0, -0.2]^\top$ m, and subsequently calculate the corresponding quadrotor position to set as the waypoint constraint. This approach enables isolated observation of orientation changes. As illustrated in \Cref{fig:strikeandstrike} (c) and the third plot in \Cref{fig:strikeandstrike} (d), the aerial manipulator successfully achieves the desired corresponding attitude at the target position.

\textbf{2) Grasp:} We design complementary grasping scenarios with two experimental configurations. In each test, an object of known mass is placed on an inclined surface, with its position tracked using the motion capture system. The aerial manipulator is tasked with planning and executing a grasping maneuver. This grasp skill requires setting an end-effector waypoint constraint, and to improve success rates, we additionally impose a velocity constraint of 0.2 m/s downward for end-effector along $\boldsymbol{z}_\mathcal{B}$ at the waypoint, while constraining velocities in $\boldsymbol{x}_\mathcal{B}$ and $\boldsymbol{y}_\mathcal{B}$ directions to zero.

To validate our framework's advantages, we again position the aerial manipulator behind the inclined surface as the initial position. The first experimental group has an initial position at (-3, 1) m with a surface inclination of approximately 42.3$^{\circ}$, while the second group starts at (-2.5, 2.5) m with an inclination of 26.7$^{\circ}$. The actual flight performance is shown in \Cref{fig:strikeandstrike} (e) and (f), demonstrating successful object grasping in both cases. \Cref{fig:strikeandstrike} (g) (first two plots) illustrates the corresponding IL-generated guide points and the resulting quadrotor and end-effector trajectories in the $xy$-plane. The third plot in \Cref{fig:strikeandstrike} (g) shows the quadrotor's grasping velocity, demonstrating rapid grasping speeds of nearly 3 m/s while confirming that the quadrotor reduces speed appropriately to satisfy the constraints. The fourth plot compares the planned versus actual thrust forces, showing that despite using a simplified model that considers only quasi-static forces and assumes fixed external forces, these assumptions remain reasonable for simple manipulation tasks. 
However, the observed discrepancies during contact phases indicate that more sophisticated contact-rich planning methods would be necessary for more complex or dynamic manipulation scenarios.

\textbf{3) Lift, Press, Wind:} These three tasks can be completed by applying one or several waypoint constraints. The lift task requires the aerial manipulator to use a hook to lift a basket, which is considered successful when the basket is lifted. We apply two waypoint constraints: one below the basket handle and another directly above it. The press task requires the end-effector to press a button, with success indicated by the button emitting red light, achieved by setting a single waypoint constraint at the button's position. The wind task involves threading a line through three hooks at known positions, with success defined as the line successfully passing through all hooks. This requires setting three end-effector waypoint constraints. The experimental results shown in \Cref{fig:multiple_tasks} (\textbf{Lift}, \textbf{Press}, \textbf{Wind}) demonstrate successful completion of all three tasks. The corresponding motion trajectory curves are presented in \Cref{fig:multiple_tasks} (a)-(c), where gray areas indicate the task execution regions.

\textbf{4) Pull, Push:} We design specific scenarios to validate the two skills independently. For the pull task, we attach a sponge to the end-effector and program the aerial manipulator to perform a table-wiping task. We set three coplanar waypoint constraints beyond the table surface and constrain the $z$-component variation of the end-effector position to zero within these points, enabling movement within a single plane. For the push task, we place a wooden block on a table surface and program the aerial manipulator to push it off the table while restricting end-effector movement to the $x$-axis direction only, with $y$ and $z$-axis position components constrained. The results shown in \Cref{fig:multiple_tasks} (\textbf{Pull}, \textbf{Push}) demonstrate successful task execution. \Cref{fig:multiple_tasks} (d) and (e) show that the gray constraint regions effectively limit motion: the pull task maintains minimal $z$-axis variation, while the push task shows consistent $y$ and $z$-axis values within the gray constraint region.

\textbf{5) Cross:} We design a narrow racing gate that requires the aerial manipulator to retract its robotic arm to successfully pass through. To increase experimental complexity and comprehensively validate our framework's capability, we extend the end-effector to the -0.2m position before and after crossing, thereby demonstrating both obstacle avoidance and manipulation capabilities. \Cref{fig:multiple_tasks} (\textbf{Cross}) shows the aerial manipulator's state during gate traversal, while \Cref{fig:multiple_tasks} (f) presents the end-effector's $z$-axis trajectory in $\mathcal{F}_\mathcal{D}$ during flight, clearly demonstrating the robotic arm's retraction-extension sequence from initial extension through retraction for gate passage to re-extension afterward.

\textbf{6) Write:} The final task we design is writing, which is comparatively complex, requiring numerous waypoint constraints while ensuring end-effector movement within the same plane. The positional constraints vary dynamically; for example, writing "T" requires initial constraint to $x$-axis movement, lifting the brush pen, then constraining to $y$-axis movement. Writing "R" requires multiple constrained movement transitions. The experimental results shown in \Cref{fig:multiple_tasks} (\textbf{Write}) demonstrate successful completion of writing the letters "TRO" using a brush pen. The actual end-effector movement can be observed in \Cref{fig:multiple_tasks} (g). We note slight oscillations at corners when writing "O," attributed to unstable connections between the end-effector and brush, as well as uneven paper surfaces. The position errors for these three experiments, shown in \Cref{fig:multiple_tasks} (h), maintain median end-effector errors within 3cm, which falls within acceptable ranges. This further validates that while our planning employs simplified models, most disturbances and unconsidered dynamic coupling effects can be resolved at the controller level, with our planner providing highly effective reference trajectories.

\section{Conclusion}\label{sec: conclusion}
This paper introduces a novel whole-body integrated motion planning framework for aerial manipulators that jointly optimizes the motions of the quadrotor and the manipulator. We propose an efficient varying ellipsoid collision avoidance method that dynamically adapts to the aerial manipulator's configuration changes, enabling safe navigation in complex environments. 
The framework introduces flexible partial waypoint constraints for precise task specification and an IL-guided optimization approach to address the limitations of standard optimization methods in handling large-attitude manipulation tasks. This IL-guided method leverages imitation learning to provide appropriate guide points, significantly improving planning success rates for challenging manipulation scenarios.
Our comprehensive experimental validation through 17 real-world experiments spanning 9 fundamental manipulation skills demonstrates the framework's excellent scalability and versatility across diverse manipulation tasks.

While our algorithm demonstrates outstanding performance across various manipulation skills, the current computational requirements prevent real-time operation. This limitation restricts the framework's ability to handle highly dynamic and complex environments where real-time replanning and adaptive adjustments are essential for safe operation.
Additionally, as evidenced by the discrepancies observed in thrust force comparisons during grasp experiments, the simplified quasi-static force models limit our ability to consider tasks involving strong interaction forces during the planning stage. Therefore, future work will focus on developing advanced dynamic models and improving computational efficiency, with the goal of enabling real-time performance in complex, contact-rich scenarios.

\bibliographystyle{IEEEtran}
\bibliography{main}

\begin{IEEEbiography}[{\includegraphics[width=1in,height=1.25in,clip,keepaspectratio]{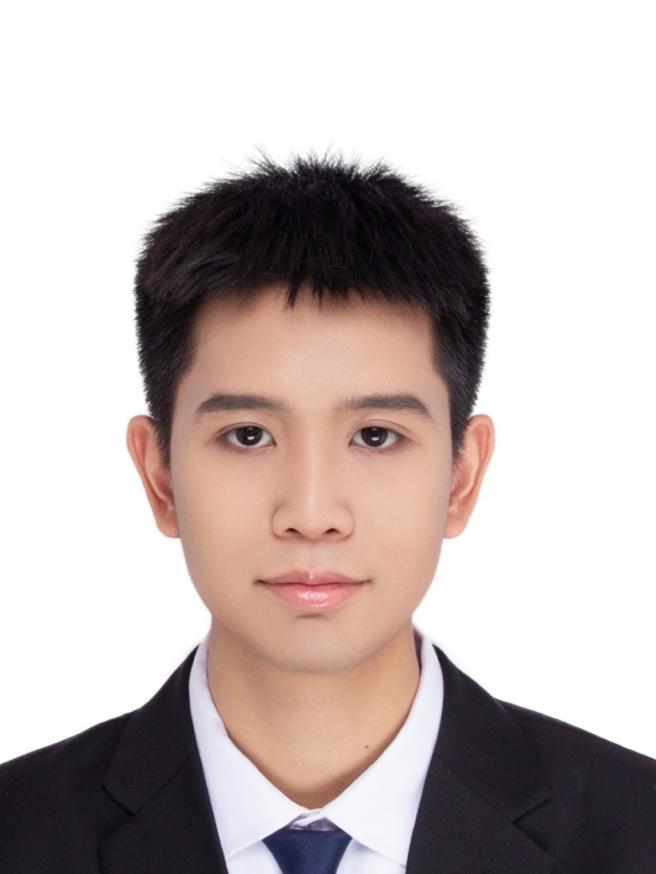}}]{Weiliang Deng} received his B.Eng. degree in Intelligence Science and Technology from Sun Yat-sen University, Shenzhen, China, in 2025. He is currently working toward the M.Phil. degree with the Department of Intelligent Systems Engineering, Sun Yat-sen University, Shenzhen, China.

His research interests include motion planning, aerial robots and robot manipulation. 
\end{IEEEbiography}
% \vspace{-1cm}

\begin{IEEEbiography}[{\includegraphics[width=1in,height=1.25in,clip,keepaspectratio]{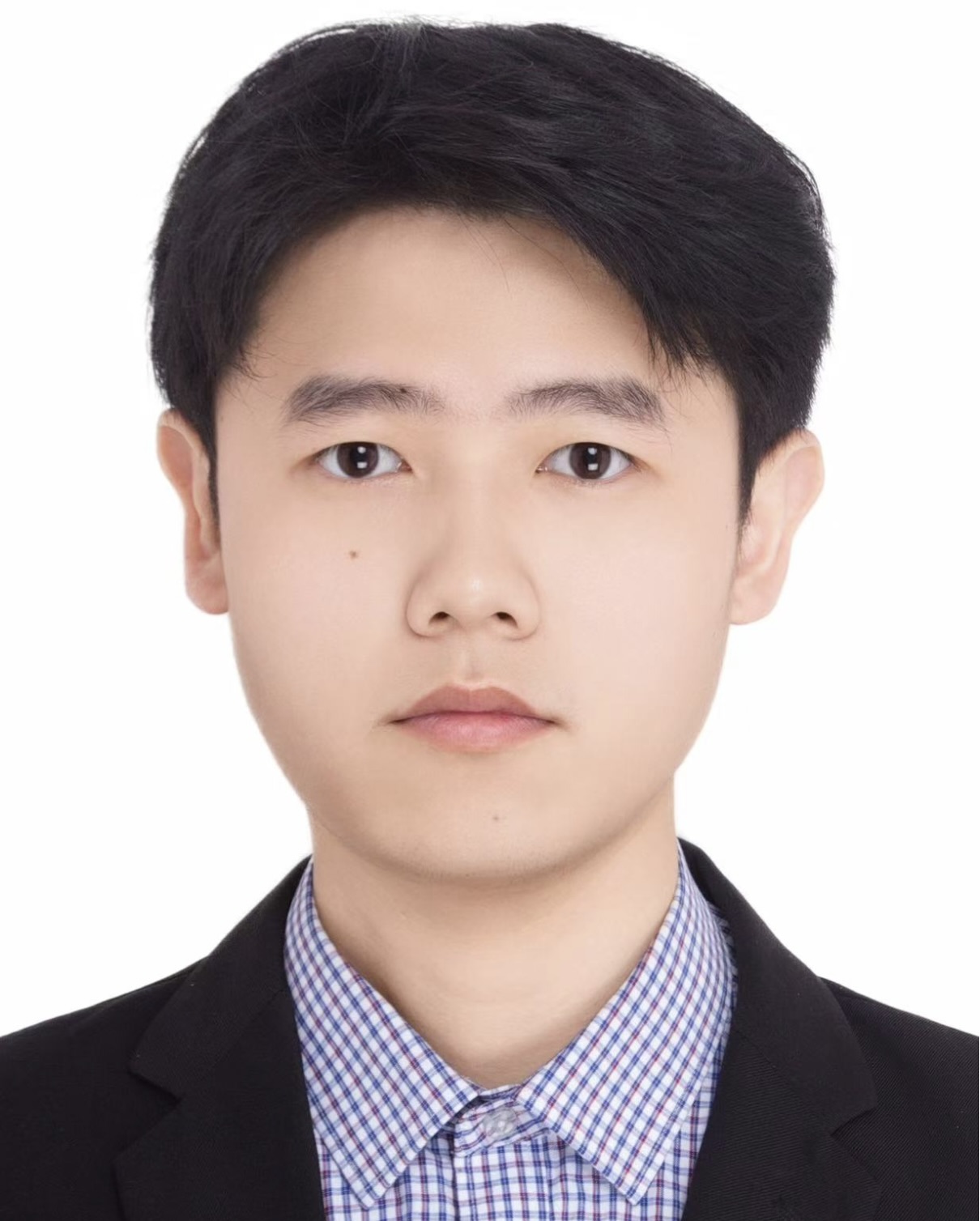}}]
{Hongming Chen} received the B.Eng. degree in Software Engineering from
the University of Electronic Science and Technology of China, Chengdu, China, in 2024. He is currently working toward the M.Phil. degree with the Department of Intelligent Systems Engineering, Sun Yat-sen University, Shenzhen, China. 

His research interests include unmanned aerial vehicles, semantic navigation, and aerial manipulation.
\end{IEEEbiography}

\begin{IEEEbiography}[{\includegraphics[width=1in,height=1.25in,clip,keepaspectratio]{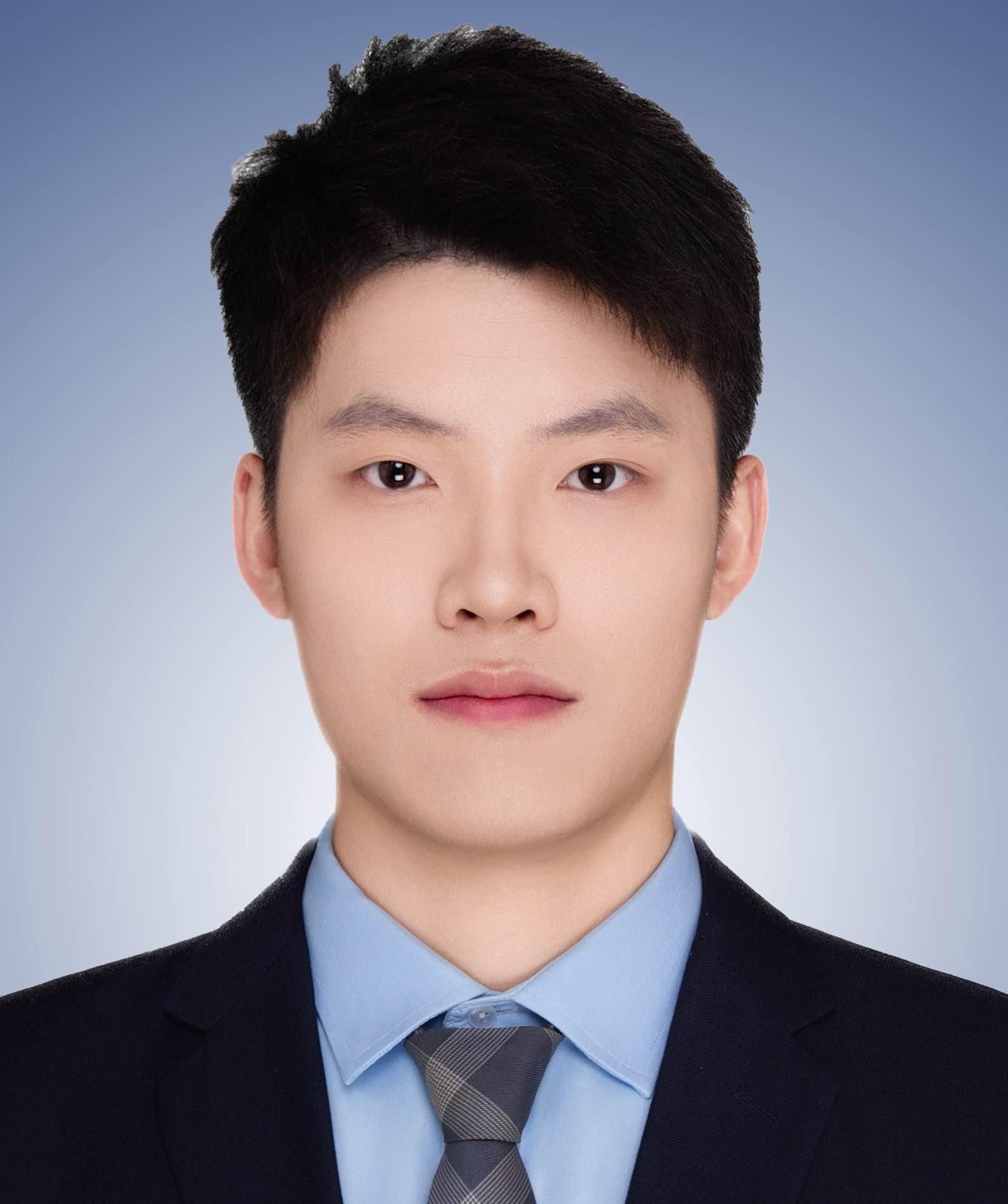}}]
{Biyu Ye} received the B.Eng. degree in Traffic Engineering from Sun Yat-sen University, Guangzhou, China, in 2024. He is currently working toward the M.S. degree in aerial vehicle design with The School of Intelligent Systems Engineering, Sun Yat-sen University.

His research interests include the perception and control of aerial robots.
\end{IEEEbiography}
% \vspace{-16.2cm}
\vfill

\newpage

\begin{IEEEbiography}[{\includegraphics[width=1in,height=1.25in,clip,keepaspectratio]{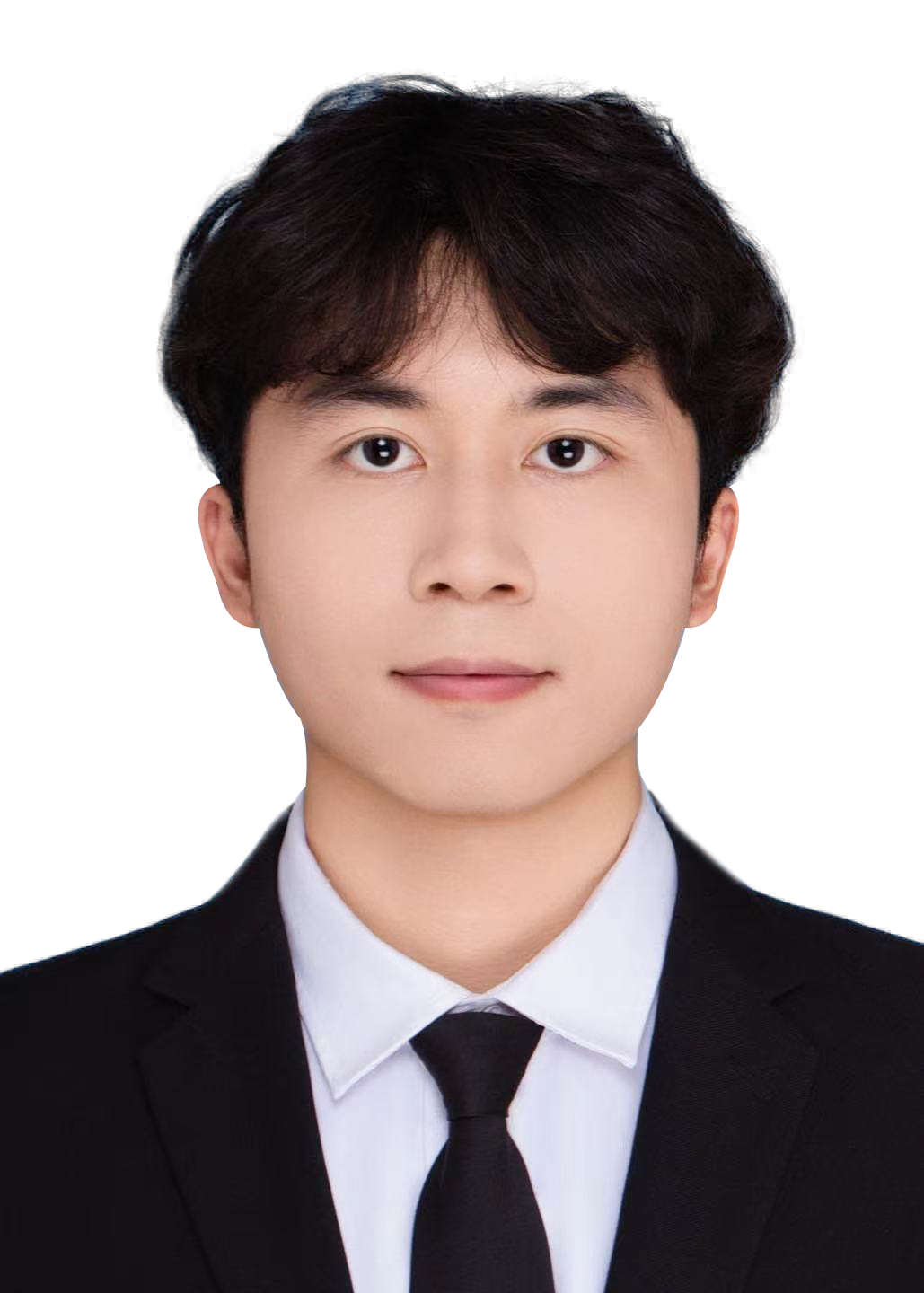}}]
{Haoran Chen} received his B.Eng. degree in Intelligence Science and Technology from Sun Yat-sen University, Shenzhen, China, in 2025. He is currently working toward the M.S. degree in the department of mechanical engineering at the National University of Singapore, Singapore. 

His research interests include aerial manipulation, and hardware design, locomotion control, and planning of legged robots.

\end{IEEEbiography}

% \newpage
% \columnbreak

\vspace{-0.3cm}

\begin{IEEEbiography}[{\includegraphics[width=1in,height=1.25in,clip,keepaspectratio]{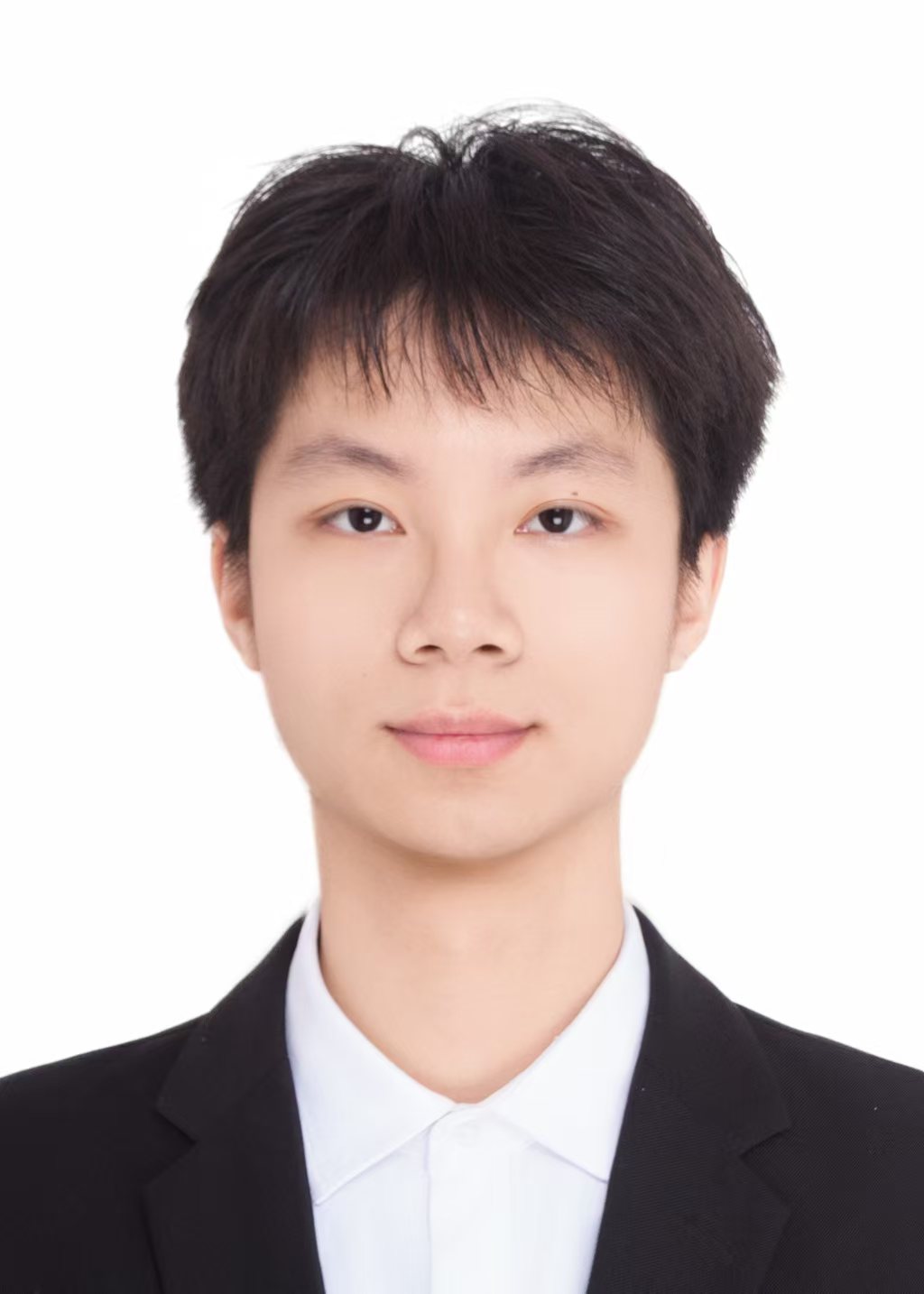}}]
{Ziliang Li} received his B.Eng. degree in Intelligence Science and Technology from Sun Yat-sen University, Shenzhen, China, in 2024. He is currently working toward the M.Phil. degree with the Department of Intelligent Systems Engineering, Sun Yat-sen University, Shenzhen, China.

His research interests include optimal control, motion planning, and aerial manipulation.
\end{IEEEbiography}

\begin{IEEEbiography}[{\includegraphics[width=1in,height=1.25in,clip,keepaspectratio]{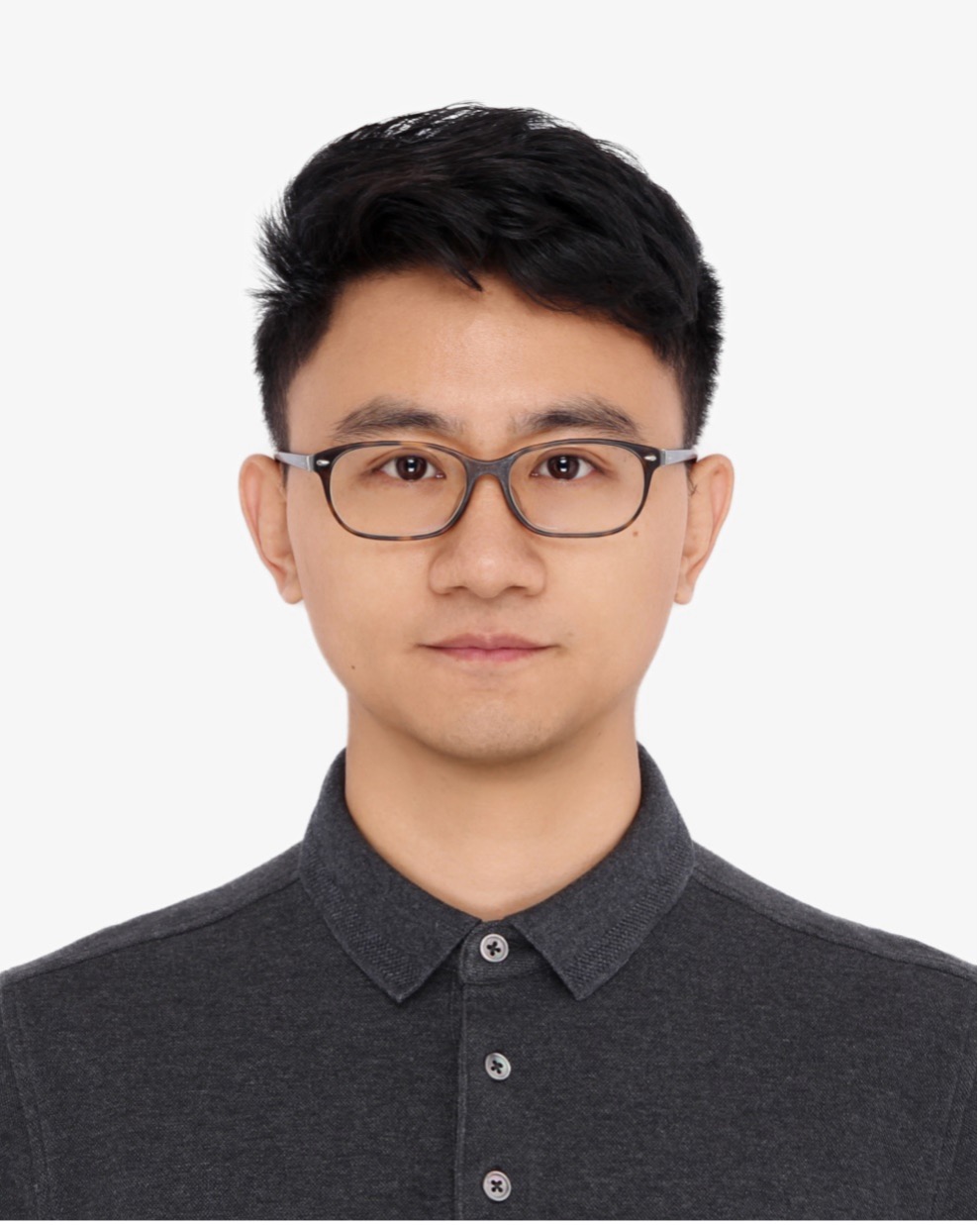}}]
{Ximin Lyu} received his B.Eng. and M.Phil. degrees in Aircraft Manufacturing from Harbin Institute of Technology, Harbin, China, in 2012 and 2014, respectively. He received his Ph.D. in Electronic and Computer Engineering from the Hong Kong University of Science and Technology, Hong Kong, China, in 2019.

In 2018, he served as a Senior Flight Control Researcher at Da Jiang Innovation (DJI) Technology Company in China. Since 2021, he has been an Associate Professor with the Department of Intelligent Systems Engineering, Sun Yat-sen University, Shenzhen, China. His research interests encompass robotics and controls, with a specific focus on UAV/UGV design, control, and planning.
\end{IEEEbiography}

\vfill

\end{document}